\documentclass[11pt,a4paper]{article}
\usepackage[hyperref]{emnlp2020}
\usepackage{times}
\usepackage{latexsym}

\usepackage{microtype}

\usepackage{graphicx}
\usepackage{multirow}
\usepackage{breqn}
\usepackage{multirow}
\usepackage{booktabs}
\usepackage{url}
\usepackage{tabularx}
\usepackage{subcaption}
\usepackage{bm}
\usepackage{comment}
\usepackage{arydshln}
\usepackage{textcomp}

\usepackage{amssymb}

\definecolor{EasyColor}{rgb}{0.141, 0.388, 0.290}
\definecolor{MediumColor}{rgb}{0.588, 0.443, 0.090}
\definecolor{HardColor}{rgb}{0.702, 0.106, 0.106}

\aclfinalcopy % Uncomment this line for the final submission
\def\aclpaperid{2574} %  Enter the acl Paper ID here

\title{Interactive Fiction Game Playing as Multi-Paragraph Reading Comprehension with Reinforcement Learning}

\author{
  Xiaoxiao Guo\thanks{\,\, Primary authors.}\\
  IBM Research\\
  \texttt{xiaoxiao.guo@ibm.com}\\
  \And
  Mo Yu$^{*}$ \\
  IBM Research\\
  \texttt{yum@us.ibm.com}\\\And
  Yupeng Gao \\
  IBM Research\\
  \texttt{yupeng.gao@ibm.com}\\
  \AND
  Murray Campbell \\
  IBM Research \\
  \texttt{mcam@us.ibm.com}\\
  \And
  Chuang Gan \\
  MIT-IBM Watson AI Lab\\
  \texttt{chuangg@ibm.com}\And
  Shiyu Chang \\
  MIT-IBM Watson AI Lab\\
  \texttt{shiyu.chang@ibm.com}\\}
  
\date{} 
\begin{document}
\maketitle
\begin{abstract}
Interactive Fiction (IF) games with real human-written natural language texts provide a new natural evaluation for language understanding techniques. 
In contrast to previous text games with mostly synthetic texts, IF games pose language understanding challenges on the human-written textual descriptions of diverse and sophisticated game worlds and language generation challenges on the action command generation from less restricted combinatorial space.
We take a novel perspective of IF game solving and re-formulate it as Multi-Passage Reading Comprehension (MPRC) tasks. Our approaches utilize the context-query attention mechanisms and the structured prediction in MPRC to efficiently generate and evaluate action outputs and apply an object-centric historical observation retrieval strategy to mitigate the partial observability of the textual observations. 
Extensive experiments on the recent IF benchmark (\emph{Jericho}) demonstrate clear advantages of our approaches achieving high winning rates and low data requirements compared to all previous approaches.\footnote{Source code is available at: \url{https://github.com/XiaoxiaoGuo/rcdqn}. }
\end{abstract}

%%%%%%%%%%%%%%%%%%%%%%%%%%%%%%%%%%%%%%%%%
\section{Introduction}

\begin{figure}[t]
\centering
\includegraphics[width=0.98\linewidth]{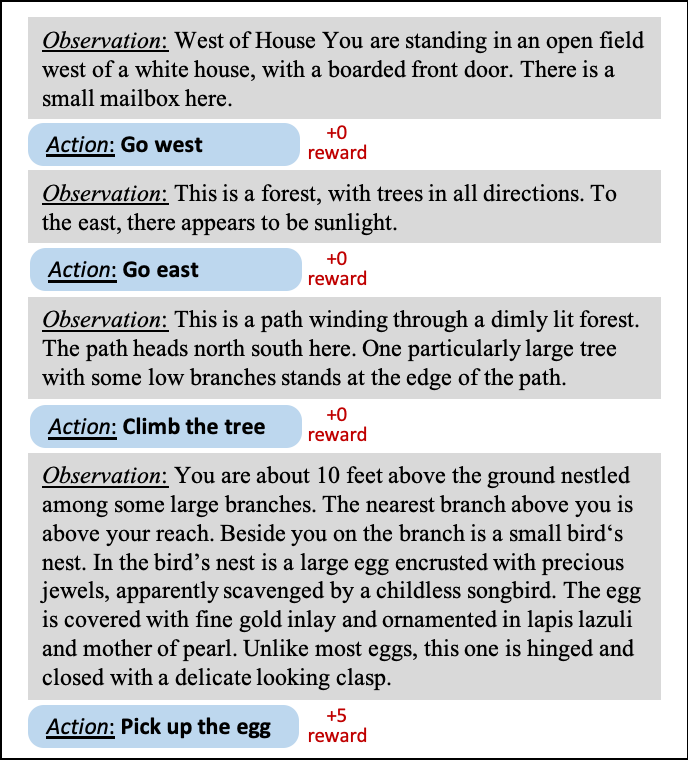}
\caption{\small Sample gameplay for the classic dungeon game \textit{Zork1}. The objective is to solve various puzzles and collect the 19 treasures to install the trophy case.  
The player receives textual observations describing the current game state and additional reward scalars encoding the game designers' objective of game progress. The player sends textual action commands to control the protagonist.
\label{fig:game}}
\end{figure}

\begin{figure*}[t]
\centering
\includegraphics[width=0.98\textwidth]{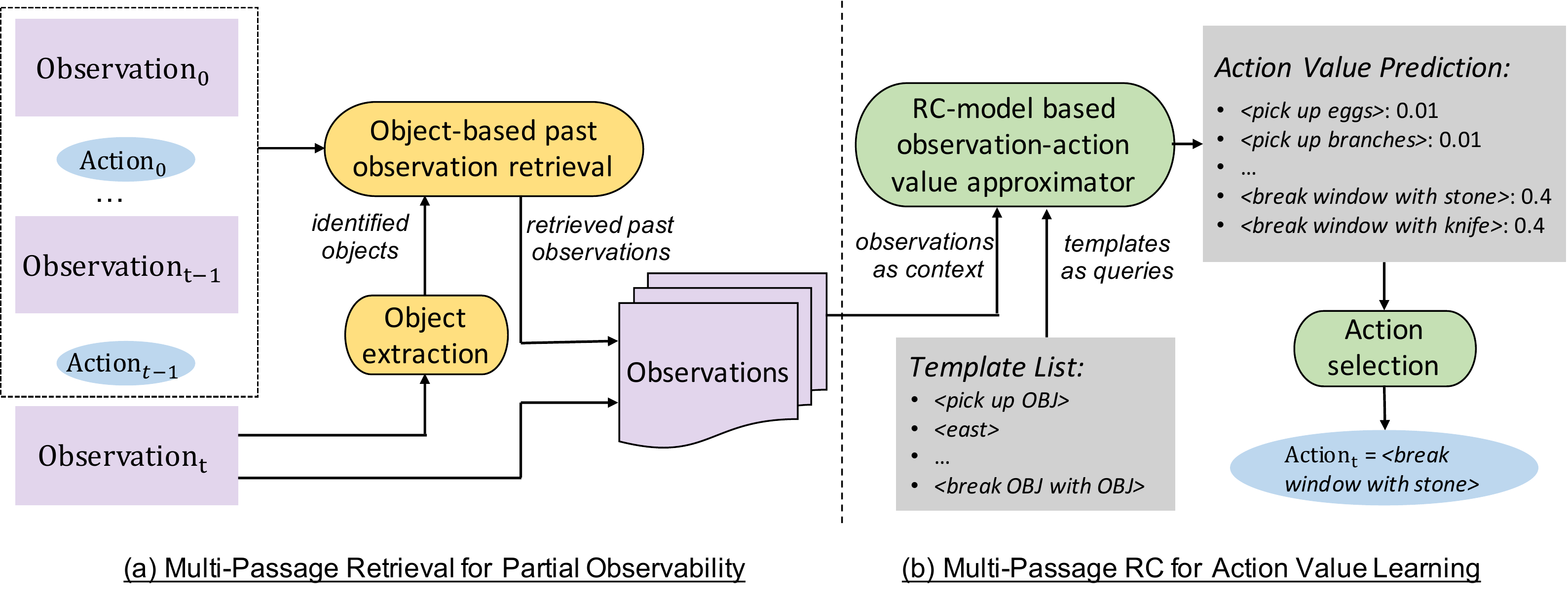}
\caption{\small Overview of our approach to solving the IF games as Multi-Paragraph Reading Comprehension (MPRC) tasks.\label{fig:framework}}
\end{figure*}

Interactive systems capable of understanding natural language and responding in the form of natural language text have high potentials in various applications. 
In pursuit of building and evaluating such systems, we study learning agents for Interactive Fiction (IF) games. IF games are world-simulating software in which players use text commands to control the protagonist and influence the world, as illustrated in Figure~\ref{fig:game}. IF gameplay agents need to simultaneously understand the game's information from a text display (\textbf{observation}) and generate natural language command (\textbf{action}) via a text input interface.  Without providing an explicit game strategy, the agents need to identify behaviors that maximize objective-encoded cumulative rewards.  

IF games composed of human-written texts (distinct from previous text games with synthetic texts) create superb new opportunities for studying and evaluating natural language understanding (NLU) techniques due to their unique characteristics. 
(1) Game designers elaborately craft on the literariness of the narrative texts to attract players when creating IF games. The resulted texts in IF games are more linguistically diverse and sophisticated than the template-generated ones in synthetic text games.
(2) The language contexts of IF games are more versatile because various designers contribute to enormous domains and genres, such as adventure, fantasy, horror, and sci-fi.
(3) The text commands to control characters are less restricted, having sizes over six orders of magnitude larger than previous text games. 
The recently introduced \textit{Jericho} benchmark provides a collection of such IF games~\cite{hausknecht2019interactive}. 

The complexity of IF games demands more sophisticated NLU techniques than those used in synthetic text games. Moreover, the task of designing IF game-play agents, intersecting NLU and reinforcement learning (RL), poses several unique challenges on the NLU techniques.
The first challenge is the difficulty of exploration in \emph{\textbf{the huge natural language action space}}.
To make RL agents learn efficiently %via trial-and-error 
without prohibitive exhaustive trials, the action estimation must generalize learned knowledge from tried actions to others. 
To this end, previous approaches, starting with a single embedding vector of the observation,
either predict the elements of actions independently~\cite{narasimhan2015language,hausknecht2019interactive}; or embed each valid action as another vector and predict action value based on the vector-space similarities~\cite{he2016deep}.
These methods do not consider the compositionality or role-differences of the action elements, or the interactions among them and the observation. Therefore, their modeling of the action values is less accurate and less data-efficient. 

The second challenge is \emph{\textbf{partial observability}}.  At each game-playing step, the agent receives a textual observation describing the locations, objects, and characters of the game world.  But the latest observation is often not a sufficient summary of the interaction history and may not provide enough information to determine the long-term effects of actions.  Previous approaches address this problem by building a representation over past observations (e.g., building a graph of objects, positions, and spatial relations)~\cite{ammanabrolu2019playing,ammanabrolu2020graph}. These methods treat the historical observations equally and summarize the information into a single vector without focusing on important contexts related to the action prediction for the current observation. Therefore, their usages of history also bring noise, and the improvement is not always significant.

We propose a novel formulation of IF game playing as Multi-Passage Reading Comprehension (MPRC) and harness MPRC techniques to solve the \emph{huge action space} and \emph{partial observability} challenges. The graphical illustration is shown in Figure~\ref{fig:framework}. 
First, the action value prediction (i.e., predicting the long-term rewards of selecting an action) is essentially
\emph{generating and scoring a compositional action structure by finding supporting evidence from the observation}.
We base on the fact that each action is an instantiation of a \textbf{template}, i.e., a verb phrase with a few placeholders of object arguments it takes~(Figure~\ref{fig:framework}b).
Then the action generation process can be viewed as extracting objects for a template's placeholders from the textual observation, based on the interaction between the template verb phrase and the relevant context of the objects in the observation.
Our approach addresses the structured prediction and interaction problems with the idea of context-question attention mechanism in RC models. 
Specifically, we treat the observation as a passage and each template verb phrase as a question. 
The filling of object placeholders in the template thus becomes an extractive QA problem that selects objects from the observation given the template. Simultaneously each action (i.e., a template with all placeholder replaced) gets its evaluation value predicted by the RC model.
Our formulation and approach better capture the fine-grained interactions between observation texts and structural actions, in contrast to previous approaches that represent the observation as a single vector and ignore the fine-grained dependency among action elements.

Second, alleviating partial observability is essentially \emph{enhancing the current observation with potentially relevant history} and \emph{predicting actions over the enhanced observation}. Our approach retrieves potentially relevant historical observations with an object-centric approach  (Figure~\ref{fig:framework}a), so that the retrieved ones are more likely to be connected to the current observation as they describe at least one shared interactable object.
Our attention mechanisms are then applied across the retrieved multiple observation texts to focus on informative contexts for action value prediction. 

We evaluated our approach on the suite of \textit{Jericho} IF games, compared to all previous approaches. Our approaches achieved or outperformed the state-of-the-art performance on \textbf{25} out of 33 games, trained with less than \textbf{one-tenth} of game interaction data used by prior art.  We also provided ablation studies on our models and retrieval strategies. 

\section{Related Work}

\paragraph{IF Game Agents.}
Previous work mainly studies the text understanding and generation in parser-based or rule-based text game tasks, such as TextWorld platform~\cite{cote2018textworld} or custom domains~\cite{narasimhan2015language,he2016deep,adhikari2020learning}. The recent platform \textit{Jericho}~\cite{hausknecht2019interactive} supports over thirty human-written IF games. 
Earlier successes in real IF games mainly rely on heuristics without learning. NAIL~\cite{hausknecht2019nail} is the state-of-the-art among these ``no-learning'' agents,  employing a series of reliable heuristics for exploring the game, interacting with objects, and building an internal representation of the game world.
With the development of learning environments like \textit{Jericho}, the RL-based agents have started to achieve dominating performance.

A critical challenge for learning-based agents is how to handle the \textbf{combinatorial action space} in IF games. LSTM-DQN~\cite{narasimhan2015language} was proposed to generate verb-object action with pre-defined sets of possible verbs and objects, but treat the selection and learning of verbs and objects independently. 
Template-DQN~\cite{hausknecht2019interactive} extended LSTM-DQN for template-based action generation, introducing one additional but still independent prediction output for the second object in the template.
Deep Reinforcement Relevance Network (DRRN)~\cite{he2016deep} was introduced for choice-based games.  Given a set of valid actions at every game state, DRRN projects each action into a hidden space that matches the current state representation vector for action selection. 
Action-Elimination Deep Q-Network (AE-DQN)~\cite{zahavy2018learn} learns to predict invalid actions in the adventure game \textit{Zork}. It eliminates invalid action for efficient policy learning via utilizing expert demonstration data.

% states 
Other techniques focus on addressing the \textbf{partial observability} in text games. 
Knowledge Graph DQN (KG-DQN)~\cite{ammanabrolu2019playing} was proposed to deal with synthetic games. The method constructs and represents the game states as knowledge graphs with objects as nodes and uses pre-trained general purposed OpenIE tool and human-written rules to extract relations between objects. KG-DQN handles the action representation following DRRN. 
KG-A2C~\cite{ammanabrolu2020graph} later extends the work for IF games, by adding information extraction heuristics to fit the complexity of the object relations in IF games and utilizing a GRU-based action generator to handle the action space.

\paragraph{Reading Comprehension Models for Question Answering.}
Given a question, reading comprehension (RC) aims to find the answer to the question based on a paragraph that may contain supporting evidence. One of the standard RC settings is extractive QA~\cite{rajpurkar2016squad,joshi2017triviaqa,kwiatkowski2019natural}, which extracts a span from the paragraph as an answer. 
Our formulation of IF game playing resembles this setting.

Many neural \emph{reader} models have been designed for RC. Specifically, for the extractive QA task, the reader models usually build question-aware passage representations via attention mechanisms~\cite{seo2016bidirectional,yu2018qanet}, and employ a pointer network to predict the start and end positions of the answer span~\cite{wang2016machine}.
Powerful pre-trained language models~\cite{peters2018deep,devlin2019bert,radford2019language} have been recently applied to enhance the encoding and attention mechanisms of the aforementioned reader models. They give performance boost but are more resource-demanding and do not suit the IF game playing task very well.

\paragraph{Reading Comprehension over Multiple Paragraphs.}
Multi-paragraph reading comprehension (MPRC) deals with the more general task of answering a question from multiple related paragraphs, where each paragraph may not necessarily support the correct answer. Our formulation becomes an MPRC setting when we enhance the state representation with historical observations and predict actions from multiple observation paragraphs.

A fundamental research problem in MPRC, which is also critical to our formulation, is to select relevant paragraphs from all the input paragraphs for the reader to focus on.
Previous approaches mainly apply traditional IR approaches like BM25~\cite{chen2017reading,joshi2017triviaqa}, or neural ranking models trained with distant supervision~\cite{wang2018r,min2019discrete}, for paragraph selection. 
Our formulation also relates to the work of evidence aggregation in MPRC~\cite{wang2017evidence,lin2018denoising}, which aims to infer the answers based on the joint of evidence pieces from multiple paragraphs.
Finally, recently some works propose the entity-centric paragraph retrieval approaches~\cite{ding2019cognitive,godbole2019multi,min2019knowledge,asai2019learning}, where paragraphs are connected if they share the same-named entities. The paragraph retrieval then becomes a traversal over such graphs via entity links. These entity-centric paragraph retrieval approaches share a similar high-level idea to our object-based history retrieval approach.
The techniques above have been applied to deal with evidence from Wikipedia, news collections, and, recently, books~\cite{mou2020frustratingly}. We are the first to extend these ideas to IF games.

%%%%%%%%%%%%%%%%%%%%%%%%%%%%%%%%%%%%%%%%%%%%%%%%%%%%%%%%%%%%%%%%%%%%%%%%%%%%%%%%%%
\section{Multi-Paragraph RC for IF Games}
\subsection{Problem Formulation}
Each IF game can be defined as a Partially Observable Markov Decision Process (POMDP), namely a 7-tuple of $\langle$ $S$, $A$, $T$, $O$, $\Omega$, $R$, $\gamma$  $\rangle$, representing the hidden game state set, the action set, the state transition function, the set of textual observations composed from vocabulary words, the textual observation function, the reward function, and the discount factor respectively.  
The game playing agent interacts with the game engine in multiple turns until the game is over or the maximum number of steps is reached.  At the $t$-th turn, the agent receives a textual observation describing the current game state $o_{t} \in O$ and sends a textual action command $a_{t} \in A$ back. The agent receives additional reward scalar $r_{t}$ which encodes the game designers' objective of game progress. Thus the task of the game playing can be formulated to generate a textual action command per step as to maximize the expected cumulative discounted rewards $\mathbf{E}\Big[ \sum_{t=0}^{\infty} \gamma^{t} r_{t}\Big]$. Value-based RL approaches learn to approximate an observation-action value function $Q(o_{t}, a_{t};\bm{\theta})$ which measures the expected cumulative rewards of taking action $a_{t}$ when observing $o_{t}$.  The agent selects action based on the action value prediction of $Q(o,a;\bm{\theta})$.   

\paragraph{Template Action Space.} 
Template action space considers actions satisfying decomposition 
in the form of $\langle \textit{verb}, \textit{arg}_{0}, \textit{arg}_{1} \rangle$. $\textit{verb}$ is an interchangeable verb phrase template with placeholders for objects and $\textit{arg}_{0}$ and $\textit{arg}_{1}$ are optional objects. For example, the action command [east], [pick up eggs] and [break window with stone] can be represented as template actions $\langle \textit{east}, \textit{none}, \textit{none} \rangle$, $\langle \textit{pick up OBJ}, \textit{eggs}, \textit{none} $ and $\langle \textit{break OBJ with OBJ}, \textit{window}, \textit{stone} \rangle$. We re-use the template library and object list from \textit{Jericho}. 
The verb phrases usually consist of several vocabulary words and each object is usually a single word. 
\begin{figure}[t]
\includegraphics[width=\linewidth]{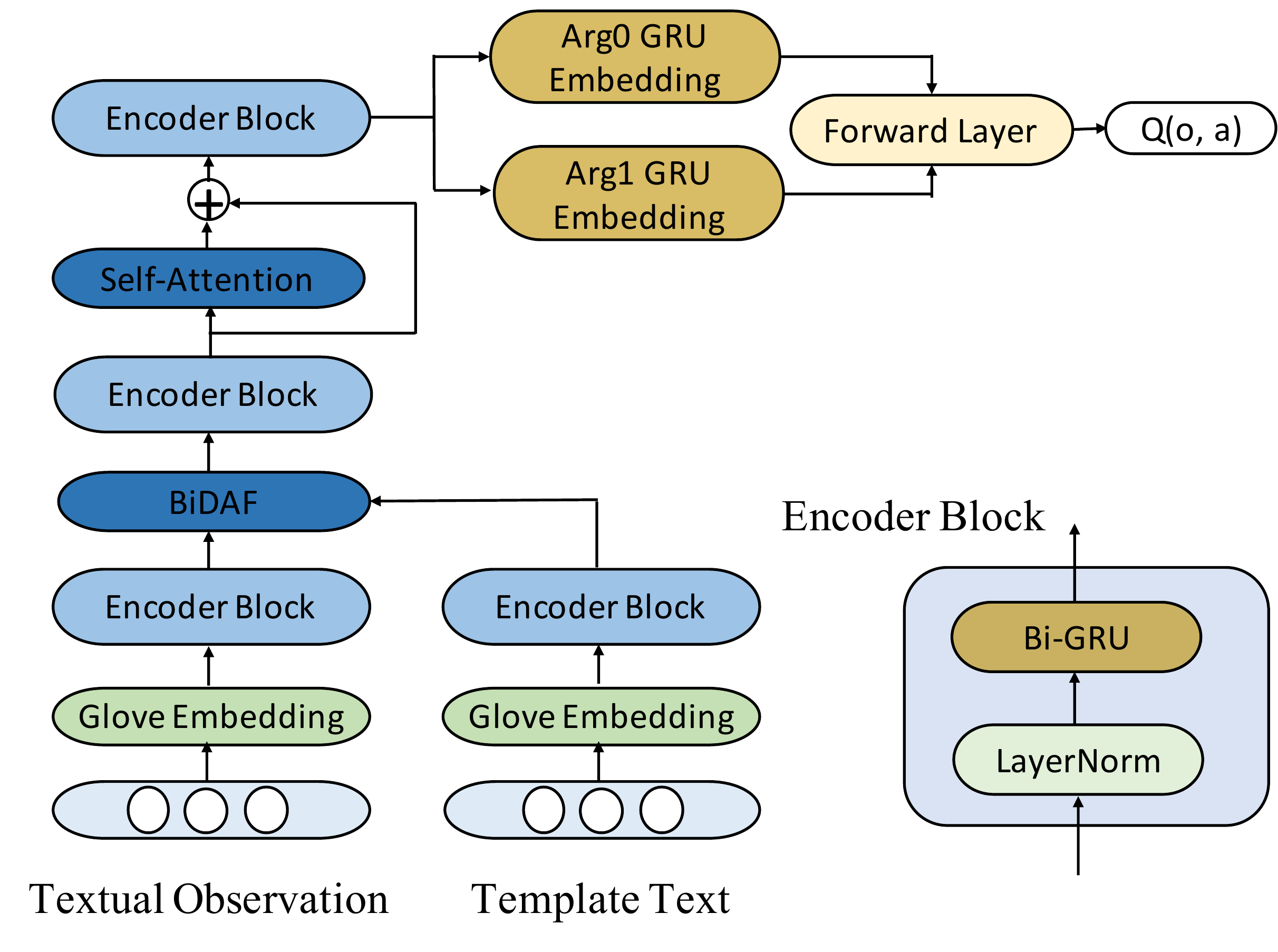}
\caption{ Our RC-based action prediction model architecture. The template text is a verb phrase with placeholders for objects, such as [pick up OBJ] and [break OBJ with OBJ]. \label{fig:model}}
\end{figure}

\subsection{RC Model for Template Actions}
We parameterize the observation-action value function $Q(o, a{=}\langle\textit{verb}, \textit{arg}_0, \textit{arg}_1\rangle ; \bm{\theta})$ by utilizing the decomposition of the template actions and context-query contextualized representation in RC. Our model treats the observation $o$ as a context in RC and the $\textit{verb}{=}(v_{1},v_{2},...,v_{k})$ component of the template actions as a query. Then a \textit{verb}-aware observation representation is derived via a RC reader model with Bidirectional Attention Flow (BiDAF)~\cite{seo2016bidirectional} and self-attention. The observation representation responding to the $\textit{arg}_{0}$ and $\textit{arg}_{1}$ words are pooled and projected to a scalar value estimate for $Q(o, a{=}\langle\textit{verb}, \textit{arg}_0, \textit{arg}_1\rangle ; \bm{\theta})$.   A high-level model architecture of our model is illustrated in Figure~\ref{fig:model}.

\paragraph{Observation and \textit{verb} Representation.}  We tokenize the observation and the \textit{verb} phrase into words, then embed these words using pre-trained GloVe embeddings~\cite{pennington2014glove}.
A shared encoder block that consists of LayerNorm~\cite{ba2016layer} and Bidirectional GRU~\cite{cho2014properties} processes the observation  and \textit{verb} word embeddings to obtain the separate observation and   \textit{verb} representation.  

\paragraph{Observation-\textit{verb} Interaction Layers.} Given the separate observation and \textit{verb} representation, we apply two attention mechanisms to compute a \textit{verb}-contextualized observation representation. We first apply BiDAF with observation as the context input and \textit{verb} as the query input.  Specifically, we denote  the processed embeddings for observation word $i$ and template word $j$ as $\bm o_i$ and $\bm t_{j}$.  The attention between the two words is then  $a_{ij}{=}\bm{w_{1}}\cdot \bm{o_{i}} + \bm{w_{2}}\cdot \bm{t_{j}} + \bm{w_{3}} \cdot (\bm{o_{i}}\otimes \bm{t_{j}})$, where $\bm{w_{1}}$, $\bm{w_{2}}$, $\bm{w_{3}}$ are learnable vectors and $\otimes$ is element-wise product.   We then compute the ``\textit{verb}2observation'' attention vector for the $i$-th observation word as  $\bm{c_{i}}{=}\sum_{j} p_{ij}\bm{t_{j}}$ with $p_{ij}{=}\exp(a_{ij})/\sum_{j}\exp(a_{ij})$.  Similarly, we compute the ``observation2\textit{verb}'' attention vector as $\bm{q}{=}\sum_{i}p_{i} \bm{o_{i}}$ with $p_{i}= \exp(\max_{j}a_{ij}) / \sum_{i}\exp(\max_{j}a_{ij})$.  We concatenate and project the output vectors as  $\bm{w_{4}}\cdot \big[\bm{o_{i}},\allowbreak \bm{c_{i}},\allowbreak \bm{o_{i}}\otimes \bm{c_{i}},\allowbreak \bm{q} \otimes \bm{c_{i}}\big]$, followed by a linear layer with leaky ReLU activation units~\cite{maas2013rectifier}.  The output vectors are processed by an encoder block.  We then apply a residual self-attention on the outputs of the encoder block. The self-attention is the same as BiDAF, but only between the observation and itself. 

\paragraph{Observation-Action Value Prediction.} We generate an action by replacing the placeholders ($\textit{arg}_{0}$ and $\textit{arg}_{1}$) in a template with  objects appearing in the observation.  The observation-action value $Q(o,a{=}\langle\textit{verb},\allowbreak \textit{arg}_{0}{=}\textit{obj}_{m}, \textit{arg}_{1}{=}\textit{obj}_{n}\rangle ;\theta)$ is achieved by processing each object's corresponding \textit{verb}-contextualized observation representation.  Specifically,  we get the indices of an \textit{obj} in the observation texts $I(\small{\textit{obj}}, o)$. When the object is a noun phrase, we take the index of its headword.\footnote{Some templates may take zero or one object. We denote the unrequired objects as \texttt{none} so that all templates take two objects. The index of the $\textit{none}$ object is for a special token. We set to the index of split token of the observation contents.}  Because the same object has different meanings when it replaces different placeholders, we apply two GRU-based embedding functions for the two placeholders, to get the object's \textit{verb}-placeholder dependent embeddings.  We derive a single vector representation $\bm{h_{\tiny \textit{arg}_{0}\textrm{=}\textit{obj}_{m}}}$ for the case that the placeholder $\textit{arg}_{0}$ is replaced by $\textit{obj}_{m}$ by mean-pooling over the \textit{verb}-placeholder dependent embeddings indexed by $I(\small{\textit{obj}_{m}}, o)$ for the corresponding placeholder $\textit{arg}_{0}$. We apply a linear transformation on the concatenated embeddings of the two placeholders to obtain the observation action value $Q(o,a){=} \bm{w}_{5}\cdot\big[\bm{h_{\tiny \textit{arg}_{0}\textrm{=}\textit{obj}_{m}}}, \bm{h_{\tiny \textit{arg}_{1}\textrm{=}\textit{obj}_{n}}} \big]$ for $a{=}\langle\textit{verb},\allowbreak \textit{arg}_{0}{=}\textit{obj}_{m}, \textit{arg}_{1}{=}\textit{obj}_{n}\rangle$. Our formulation avoids the repeated computation overhead among different actions with a shared template verb phrase. 

\begin{table*}[t!]
\centering%
\small
\begin{tabular}{lccc}
\toprule
\small \bf Agents & \bf Action strategy & \bf State strategy &  \bf Interaction data \\
\midrule
\small TDQN  & \multicolumn{1}{p{5.2cm}}{Independent selection of template and the two objects} & $\textit{None}$ & \small 1M \\ 
\midrule
\small DRRN & \multicolumn{1}{p{5.2cm}}{Action as a word sequence without distinguishing the roles of verbs and objects} & \small $\textit{None}$ & \small 1M \\ 
\midrule
\small KG-A2C & \multicolumn{1}{p{5.2cm}}{Recurrent neural decoder that selects the template and objects in a fixed order}  & \multicolumn{1}{p{5.5cm}}{Object graph from historical observations based on OpenIE and human-written rules} & \small 1.6M\\
\midrule
\small Ours  & \multicolumn{1}{p{5.2cm}}{Observation-template representation for object-centric value prediction} & \multicolumn{1}{p{5.5cm}}{Object-based history observation retrieval} & \small 0.1M \\
\bottomrule
\end{tabular}
\caption{\small Summary of the main technical differences between our agent and the baselines. All agents use DQN to update the model parameters except KG-A2C uses A2C. All agents use the same handicaps.\label{tab:summary}}
\label{tab:method_comparison}
\end{table*}

\subsection{Multi-Paragraph Retrieval Method for Partial Observability}
The observation at the current step sometimes does not have full-textual evidence to support action selection and value estimation, due to the inherent partial observability of IF games. For example, when repeatedly attacking a troll with a sword, the player needs to know the effect or feedback of the last attack to determine if an extra attack is necessary.  It is thus important for an agent to efficiently utilize historical observations to better support action value prediction.  In our RC-based action prediction model, the historical observation utilization can be formulated as selecting evidential observation paragraphs in history, and predicting the action values from multiple selected observations, namely a Multiple-Paragraph Reading Comprehension (MPRC) problem. We propose to retrieve past observations with an object-centric approach. 

\paragraph{Past Observation Retrieval.}  
Multiple past observations may share objects with the current observation, and it is computationally expensive and unnecessary to retrieve all of such observations. The utility of past observations associated with each object is often time-sensitive in that new observations may entirely or partially invalidate old observations. We thus propose a time-sensitive strategy for retrieving past observations.  Specifically, given the detected objects from the current observation, we retrieve the most recent $K$ observations with at least one shared object. 
The $K$ retrieved observations are sorted by time steps and concatenated to the current observation. The observations from different time steps are separated by a special token. Our RC-based action prediction model treats the concatenated observations as the observation inputs, and no other parts are changed. We use the notation $o_{t}$ to represent the current observation and the extended current observation interchangeably.        

\subsection{Training Loss}
We apply the Deep Q-Network (DQN)~\cite{mnih2015human} to update the parameters $\bm{\theta}$ of our RC-based action prediction model. The loss function is:
\begin{dmath*}
\mathcal{L}(\theta) = \mathbf{E}_{(o_{t}, a_{t}, r_{t} , o_{t+1})\sim \rho(\mathcal{D})} \Big[ ||Q(o_{t}, a_{t}; \theta) - (r_{t} + \gamma \max_{b} Q(o_{t+1}, b; \theta^{-})) || \Big]
\end{dmath*}
where $\mathcal{D}$ is the experience replay consisting of recent gameplay transition records and $\rho$ is a distribution over the transitions defined by a sampling strategy.

\paragraph{Prioritized Trajectories.} 
The distribution $\rho$ has a decent impact on DQN performance.  Previous work samples transition tuples with immediate positive rewards more frequently to speed up learning~\cite{narasimhan2015language,hausknecht2019interactive}.  We observe that this heuristic is often insufficient. Some transitions with zero immediate rewards or even negative rewards are also indispensable in recovering well-performed trajectories. We thus extend the strategy from transition level to trajectory level. We prioritize transitions from trajectories that outperform the exponential moving average score of recent  trajectories. 

\begin{table*}[t]
\small
\centering
\begin{tabular}{lccccccc}
\toprule
            &   \multicolumn{2}{c}{\bf Human}  &    \multicolumn{3}{c}{\bf Baselines}     & \multicolumn{2}{c}{\bf Ours} \\
\bf Game        & \bf Max & \bf Walkthrough-100   & \bf TDQN  & \bf DRRN  & \bf KG-A2C & \bf MPRC-DQN & \bf RC-DQN \\ 
\midrule
\textcolor{EasyColor}{905}         & 1  & 1   & 	\textbf{0}     & 	\textbf{0}     & 	\textbf{0}     & 	\textbf{0}     & 	\textbf{0}\\ 
\textcolor{EasyColor}{acorncourt}  & 30 & 30   & 1.6   & 	\textbf{10}    & 0.3   & 	\textbf{10.0}    & 	\textbf{10.0} \\ 
\textcolor{MediumColor}{advent}      & 350& 113  & 36    & 36    & 36    & 	\textbf{63.9}     & 36\\ 
\textcolor{EasyColor}{adventureland} & 100& 42 & 0     & 20.6  & 0     &	\textbf{24.2}   &21.7\\ 
\textcolor{EasyColor}{afflicted}   & 75 & 75   & 1.3   & 2.6   & --    &	\textbf{8.0}   &	\textbf{8.0} \\ 
\textcolor{HardColor}{anchor}      & 100 & 11  & 	\textbf{0}     & 	\textbf{0}     & 	\textbf{0}     & 	\textbf{0}     & 	\textbf{0}\\ 
\textcolor{EasyColor}{awaken}      & 50 & 50  & 	\textbf{0}     & 	\textbf{0}     & 	\textbf{0}     & 	\textbf{0}     & 	\textbf{0}\\ 
\textcolor{MediumColor}{balances}    & 51& 30    & 4.8   & 	\textbf{10}    & 	\textbf{10}    & 	\textbf{10}    & 	\textbf{10}\\ 
\textcolor{MediumColor}{deephome}    & 300 & 83  & 	\textbf{1}     & 	\textbf{1}     & 	\textbf{1}     & 	\textbf{1}     & 	\textbf{1}\\ 
\textcolor{EasyColor}{detective}   & 360 & 350  & 169   & 197.8 & 207.9 & 	\textbf{317.7} & 291.3\\ 
\textcolor{EasyColor}{dragon}      & 25 & 25   & -5.3  & -3.5  & 0     & 0.04     &	\textbf{4.84}\\ % 
\textcolor{HardColor}{enchanter}   & 400 & 125  & 8.6   & 	\textbf{20}    & 12.1  & 	\textbf{20.0}    & 	\textbf{20.0} \\ 
\textcolor{MediumColor}{gold}        & 100& 30   & 	\textbf{4.1}   & 0     & --    & 0     & 0\\ 
\textcolor{EasyColor}{inhumane}    & 90 & 70   & 0.7   & 0     & 	\textbf{3}     & 0     & 0\\ 
\textcolor{MediumColor}{jewel}       & 90 & 24   & 0     & 1.6   & 1.8   & 	\textbf{4.46}   & 2.0 \\ 
\textcolor{MediumColor}{karn}        & 170 & 40  & 0.7   & 2.1   & 0     & 	\textbf{10.0}  &	\textbf{10.0}\\ 
\textcolor{EasyColor}{library}     & 30 & 30   & 6.3   & 17    & 14.3  & 17.7  & 	\textbf{18.1} \\  
\textcolor{MediumColor}{ludicorp}    & 150 & 37  & 6     & 13.8  & 17.8 & 	\textbf{19.7}  &  17.0  \\ 
\textcolor{EasyColor}{moonlit}     & 1  & 1   & 	\textbf{0}     & 	\textbf{0}     & 	\textbf{0}     & 	\textbf{0}     & 	\textbf{0}\\ 
\textcolor{EasyColor}{omniquest}   & 50 & 50   & 	\textbf{16.8}  & 10    & 3     & 10.0  & 10.0\\ 
\textcolor{EasyColor}{pentari}     & 70 & 60   & 17.4  & 27.2  & 	\textbf{50.7}  & 44.4  & 43.8\\ 
\textcolor{EasyColor}{reverb}      & 50 & 50   & 0.3   & 	\textbf{8.2}    & --    & 2.0   & 2.0\\ 
\textcolor{EasyColor}{snacktime}   & 50 & 50   & 	\textbf{9.7}   & 0     & 0     & 0     & 0\\ 
\textcolor{HardColor}{sorcerer}    & 400 & 150  & 5     & 20.8  & 5.8   &  	\textbf{38.6}   & 38.3 \\ 
\textcolor{HardColor}{spellbrkr}   & 600 & 160  & 18.7  & 	\textbf{37.8}  & 21.3  & 25    & 25\\ %
\textcolor{HardColor}{spirit}      & 250 & 8  & 0.6   & 0.8   & 1.3   &  3.8  & 	\textbf{5.2} \\ 
\textcolor{EasyColor}{temple}      & 35 & 20   & 7.9   & 7.4   & 7.6   & 	\textbf{8.0}     & 	\textbf{8.0}\\ 
\textcolor{HardColor}{tryst205}    & 350 & 50  & 0     & 9.6   & --    & 	\textbf{10.0}   & 	\textbf{10.0}\\ 
\textcolor{MediumColor}{yomomma}     & 35 & 34   & 0     & 0.4   & --    & 	\textbf{1.0}     & 	\textbf{1.0}\\ 
\textcolor{MediumColor}{zenon}       & 20 & 20   & 0     & 0     & 	\textbf{3.9}   & 0     & 0\\
\textcolor{MediumColor}{zork1}       & 350 & 102  & 9.9   & 32.6  & 34    & 38.3  & 	\textbf{38.8}\\
\textcolor{MediumColor}{zork3}       & 7  & 3$^{a}$   & 0     & 0.5   & 0.1   & 	\textbf{3.63}   & 	2.83\\
\textcolor{EasyColor}{ztuu}        & 100$^{b}$ & 100  & 4.9   & 21.6  & 9.2   & 	\textbf{85.4}  & 79.1\\ 
\midrule
\emph{Winning} & &   & 24\%/8 & 30\%/10 & 27\%/9 & \bf 64\%/21 & 52\%/17  \\
\emph{percentage / counts} & & & & & & \multicolumn{2}{c}{76\%/25}\\
\bottomrule
\end{tabular}
\caption{\label{tab:results}\small{Average game scores on Jericho benchmark games. The best performing agent score per game is \textbf{in bold}.
\newline
The \textit{Winning percentage / counts} row computes the percentage / counts of games that the corresponding agent is best. 
The scores of baselines are from their papers. The missing scores are represented as ``--'', for which games KG-A2C skipped. 
 We also added the 100-step results from a human-written game-playing walkthrough, as a reference of human-level scores. We denote the difficulty levels of the games defined in the original Jericho paper with colors in their names -- possible (i.e., easy or normal) games in \textcolor{EasyColor}{green} color, difficult games in \textcolor{MediumColor}{tan} and extreme games in \textcolor{HardColor}{red}. Best seen in color.}
\newline \small{$^a$ \textit{Zork3} walkthrough does not maximize the score in the first 100 steps but explores more.}  \small{$^b$ Our agent discovers some unbounded reward loops in the game \textit{Ztuu}. }}
\end{table*}

\section{Experiments}

We evaluate our proposed methods on the suite of Jericho supported games. We compared to all previous baselines that include recent methods addressing the huge action space and partial observability challenges. 
 
\subsection{Setup} 

\paragraph{\textit{Jericho} Handicaps and  Configuration. } 
The handicaps used by our methods are the same as other baselines. First, we use the Jericho API to check if an action is valid with game-specific templates. Second, we augmented the observation with the textual feedback returned by the command [$\textit{inventory}$] and [$\textit{look}$]. Previous work also included the last action or game score as additional inputs. Our model discarded these two types of inputs as we did not observe a significant difference by our model.
The maximum game step number is set to 100 following baselines.

\paragraph{Implementation Details.}
We apply spaCy\footnote{\url{https://spacy.io}} to tokenize the observations and detect the objects in the observations. We use the $100$-dimensional GloVe embeddings as fixed word embeddings. The out-of-vocabulary words are mapped to a randomly initialized embedding.  The dimension of Bi-GRU hidden states is 128. We set the observation representation dimension to be $128$ throughout the model. The history retrieval window $K$ is 2. For DQN configuration, we use the $\epsilon$-greedy strategy for exploration, annealing $\epsilon$ from $1.0$ to $0.05$. $\gamma$ is $0.98$. We use Adam to update the weights with $10^{-4}$ learning rate. Other parameters are set to their default values. More details of the Reproducibility Checklist is in Appendix~\ref{app:checklist}.

\paragraph{Baselines.} 
We compare with all the public results on the Jericho suite, namely TDQN~\cite{hausknecht2019interactive}, DRRN~\cite{he2016deep}, and KG-A2C~\cite{ammanabrolu2020graph}. As discussed, our approaches differ from them mainly in the strategies of handling the large action space and partial observability of IF games. We summarize these main technical differences in Table~\ref{tab:summary}. In summary, all previous agents predict actions conditioned on a single vector representation of the whole observation texts. Thus they do not exploit the fine-grained interplay among the template components and the observations. Our approach addresses this problem by formulating action prediction as an RC task, better utilizing the rich textual observations with deeper language understanding.

\begin{table*}[t!]
\small
\centering
\begin{tabular}{lcccccc}
\toprule
\bf Game  & \bf Template Action  &\bf Avg. Steps  & \bf Dialog & \bf Darkness & \bf Nonstandard  & \bf Inventory  \\ 
& \bf Space ($\times 10^6$)  & \bf Per Reward & \bf Actions & \bf Limit \\
\midrule
\textcolor{MediumColor}{advent} & 107& 7& &\checkmark &\checkmark &\checkmark\\
\textcolor{EasyColor}{detective} & 19  & 2 &&&& \\
\textcolor{MediumColor}{karn} & 63 & 17 & \checkmark & \checkmark \\
\textcolor{MediumColor}{ludicorp} & 45  & 4 & & & \checkmark & \checkmark  \\
\textcolor{EasyColor}{pentari} & 32  & 5 & & & \checkmark & \\
\textcolor{HardColor}{spirit} & 195  & 21 & \checkmark & \checkmark & \checkmark & \checkmark \\
\textcolor{MediumColor}{zork3} & 67  & 39 & \checkmark & \checkmark & & \checkmark\\
\bottomrule
\end{tabular}
\caption{\small \label{tab:difficulty_results} Difficulty levels and characteristics of games on which our approach achieves the most considerable improvement. \emph{Dialog} indicates that it is necessary to speak with another character. \emph{Darkness} indicates that accessing some dark areas requires a light source. \emph{Nonstandard Actions} refers to actions with words not in an English dictionary. \emph{Inventory Limit} restricts the number of items carried by the player. Please refer to~\cite{hausknecht2019interactive} for more comprehensive definitions.}
\end{table*}

\paragraph{Training Sample Efficiency.} 
We update our models for $100,000$ times. Our agents interact with the environment one step per update, resulting in a total of $0.1$M 
environment interaction data. Compared to the other agents, such as KG-A2C ($1.6$M), TDQN ($1$M), and DRRN ($1$M), our environment interaction data is significantly smaller.  

\subsection{Overall Performance}
We summarize the performance of our Multi-Paragraph Reading Comprehension DQN (MPRC-DQN) agent and baselines in Table~\ref{tab:results}. 
Of the 33 IF games, our MPRC-DQN achieved or improved the state of the art performance on 21 games (i.e., a winning rate of $64$\%). The best performing baseline (DRRN) achieved the state-of-the-art performance on only ten games, corresponding to the winning rate of $30$\%, lower than half of ours. Note that all the methods achieved the same initial scores on five games, namely \emph{905}, \emph{anchor}, \emph{awaken}, \emph{deephome}, and \emph{moonlit}. Apart from these five games, our MPRC-DQN achieved more than three times wins. Our MPRC-DQN achieved significant improvement on some games, such as \emph{adventureland}, \emph{afflicted}, \emph{detective}, etc. Appendix~\ref{app:traj} shows some  game playing trajectories.

We include the performance of an RC-DQN agent, which implements our RC-based action prediction model but only takes the current observations as inputs. It also outperformed the baselines by a large margin. After we consider the RC-DQN agent, our MPRC-DQN still has the highest winning percentage, indicating that our RC-based action prediction model has a significant impact on the performance improvement of our MPRC-DQN and the improvement from the multi-passage retrieval is also unneglectable.
Moreover, compared to RC-DQN, our MPRC-DQN has another advantage of faster convergence.
The learning curves of our MPRC-DQN and RC-DQN agents on various games are in Appendix~\ref{app:curves}. 

Finally, our approaches, overall, achieve the new state-of-the-art on 25 games (i.e., a winning rate of $76$\%), giving a significant advance in the field of IF game playing.

\begin{table}[h!]
\small
\centering
\begin{tabular}{lccc}
\toprule
\bf Competitors        & \bf Win & \bf Draw   & \bf Lose\\ 
\midrule
MPRC-DQN v.s. TDQN & 23 & 6 & 4\\
MPRC-DQN v.s. DRRN & 18 & 13 & 2\\
MPRC-DQN v.s. KG-A2C & 18 & 7 & 3\\
\bottomrule
\end{tabular}
\caption{\small \label{tab:pair_results} Pairwise comparison between our MPRC-DQN versus each baseline.}
\end{table}

\paragraph{Pairwise Competition.}
To better understand the performance difference between our approach and each of the baselines, we adopt a direct one-to-one comparison metric based on the results from Table~\ref{tab:results}.
Our approach has a high winning rate when competing with any of the baselines, summarized in Table~~\ref{tab:pair_results}. All the baselines have a rare chance to beat us on games. DRRN gives a higher chance of draw-games when competing with ours.

\paragraph{Human-Machine Gap.}
We additionally compare IF gameplay agents to human players to better understand the improvement significance and the potential improvement upper-bound. We measure each agent's game progress as the macro-average of the normalized agent-to-human game score ratios, capped at 100\%.   The progress of our MPRC-DQN is 28.5\%, while the best performing baseline DRRN is 17.8\%, showing that our agent's improvement is significant even in the realm of human players. Nevertheless, there is a vast gap between the learning agents and human players. The gap indicates IF games can be a good benchmark for the development of natural language understanding techniques.

\begin{figure*}[!t]
\begin{subfigure}{.32\textwidth}
  \centering
  \includegraphics[width=.95\linewidth]{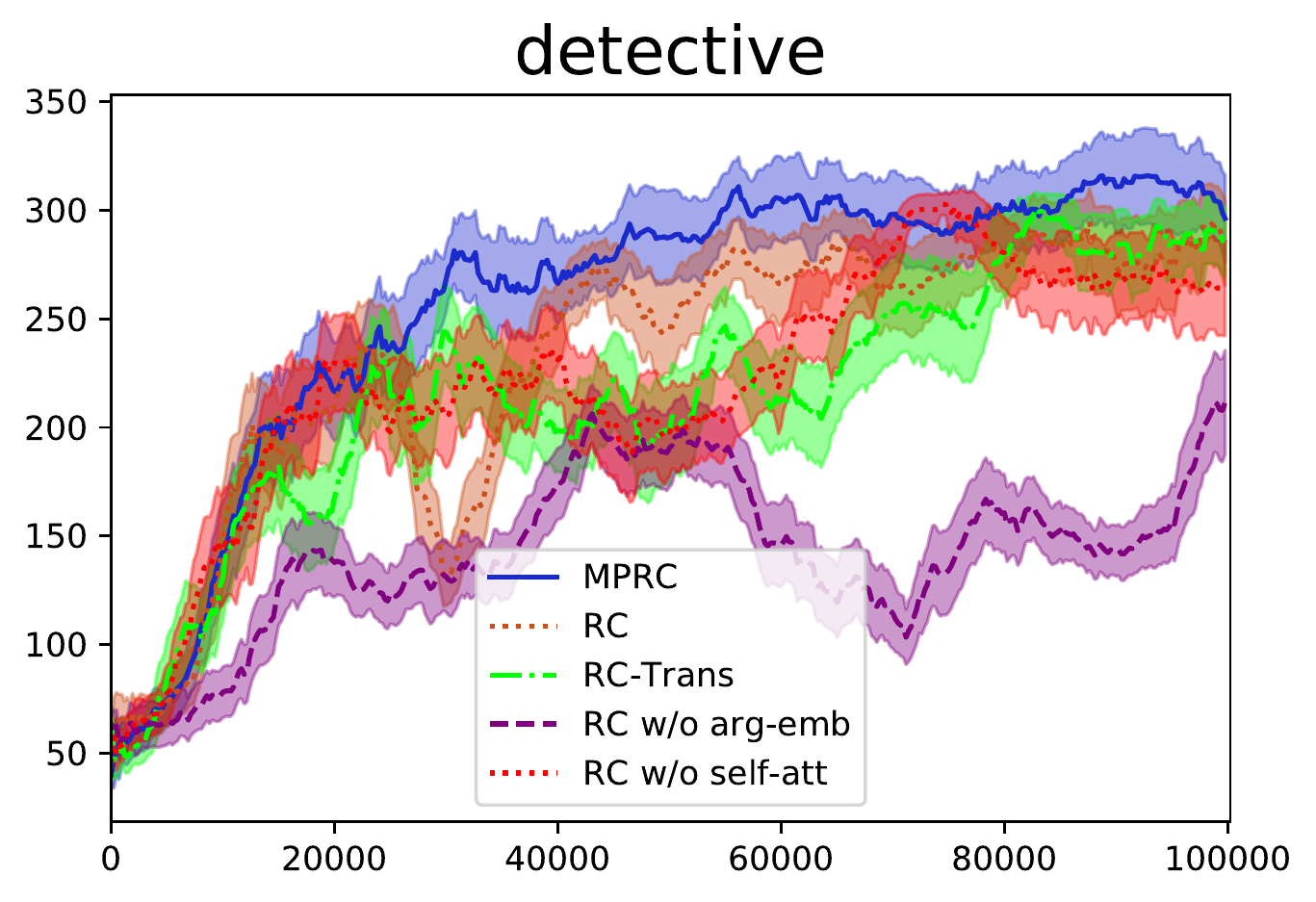}
\end{subfigure}%
\begin{subfigure}{.32\textwidth}
  \centering
  \includegraphics[width=0.95\linewidth]{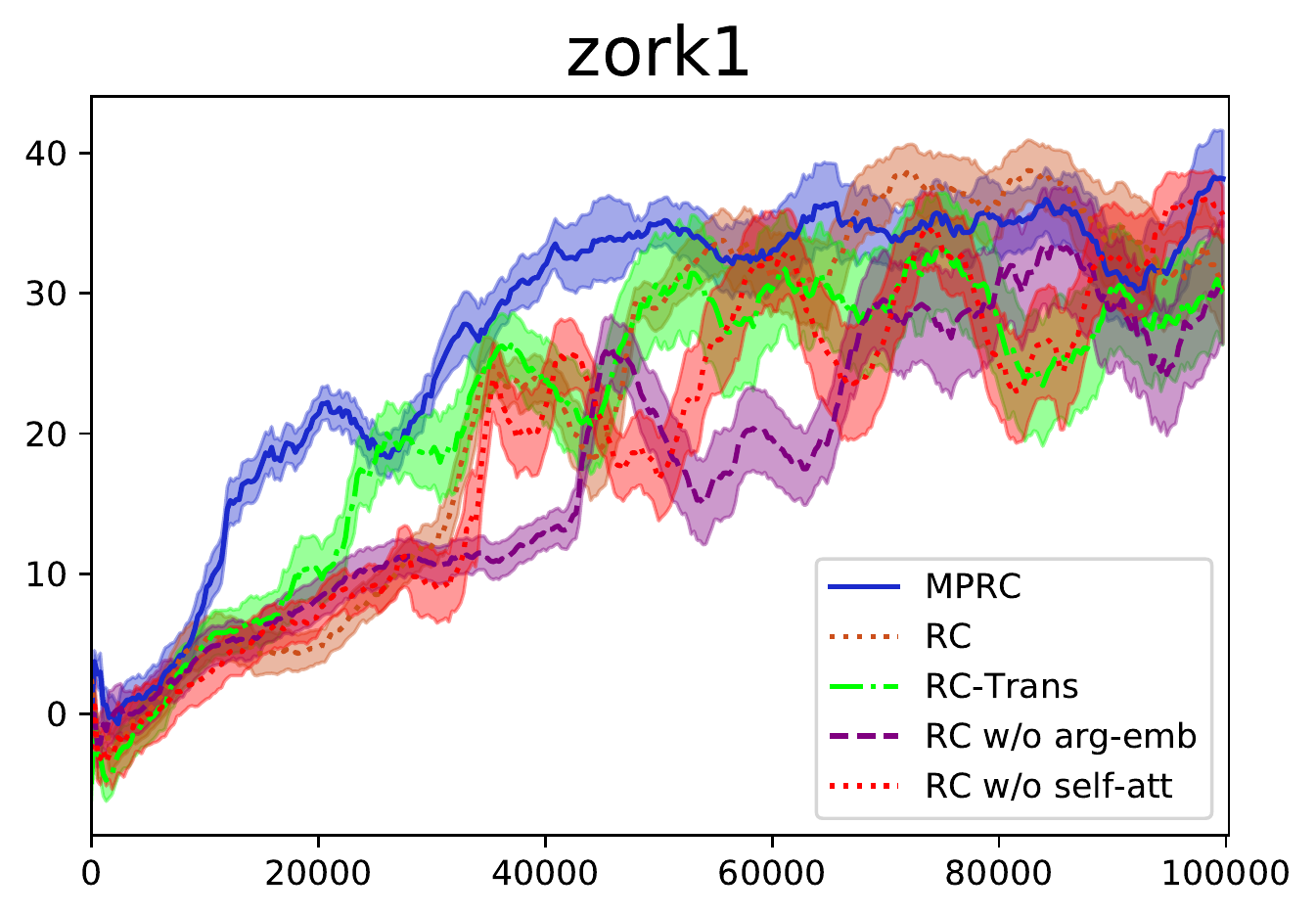}
\end{subfigure}%
\begin{subfigure}{.32\textwidth}
  \centering
  \includegraphics[width=.95\linewidth]{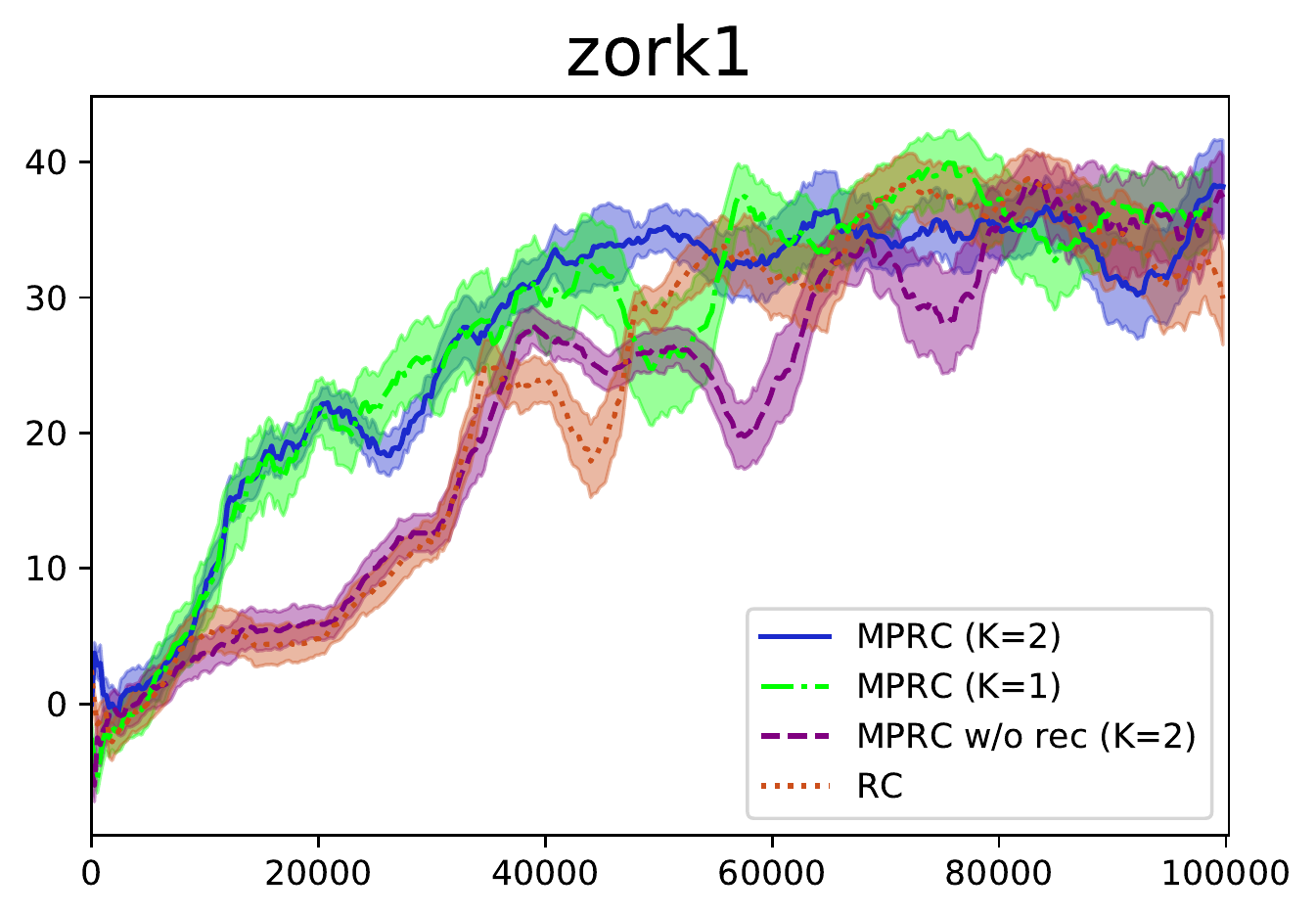}
\end{subfigure}
\caption{\small Learning curves for ablative studies. \textbf{(left)} Model ablative studies on the game \textit{Detective}. \textbf{(middle)} Model ablative studies on \textit{Zork1}.  \textbf{(right)} Retrieval strategy study on \textit{Zork1}. Best seen in color. \label{fig:ab} }
\label{fig:fig}
\end{figure*}

\paragraph{Difficulty Levels of Games.}
Jericho categorizes the supported games into three difficulty levels, namely possible games, difficult games, and extreme games, based on the characteristics of the game dynamics, such as the action space size, the length of the game, and the average number of steps to receive a non-zero reward. Our approach improves over prior art on seven of the sixteen possible games, seven of the eleven difficult games, and three of the six extreme games in Table~\ref{tab:results}. It shows that the strategies of our method are generally beneficial for any difficulty levels of game dynamics. Table~\ref{tab:difficulty_results} summarizes the characteristics of the seven games in which our method improves the most,  i.e., larger than 15\% of the game progress in the first 100 steps.\footnote{We ignore \emph{ztuu} due to the infinite reward loops.}
First, these mostly improved games have medium action space sizes, and it is an advantageous setting for our methods where modeling the template-object-observation interactions is effective.
Second, our approach improves most on games with a reasonably high degree of reward sparsity, such as \emph{karn}, \emph{spirit}, and \emph{zork3}, indicating that our RC-based value function formulation helps in optimization and mitigates the reward sparsity. 
Finally, we remark that these game difficulty levels are not directly categorized based on natural language-related characteristics, such as text comprehension and puzzle-solving difficulties. Future studies on additional game categories based on those natural language-related characteristics would shed light on related improvements. 

\subsection{Ablative Studies}

\paragraph{RC-model Design.}
The overall results show that our RC-model plays a critical role in performance improvement. We compare our RC-model to some alternative models as ablative studies.
We consider three alternatives, namely (1) our RC-model without the self-attention component (\texttt{w/o self-att}), (2) without the argument-specific embedding (\texttt{w/o arg-emb}) and (3) our RC-model with Transformer-based block encoder (\texttt{RC-Trans}) following QANet~\cite{yu2018qanet}. Detailed architecture is in Appendix~\ref{app:checklist}.

The learning curves for different RC-models are in Figure~\ref{fig:ab}~(left/middle). The RC-models without either self-attention or argument-specific embedding degenerate, and the argument-specific embedding has a greater impact. The Transformer-based encoder block sometimes learns faster than Bi-GRU at the early learning stage. It achieved a comparable final performance, even with much greater computational resource requirements.

\paragraph{Retrieval Strategy.}
We compare with history retrieval strategies with different history sizes ($K$) and pure recency-based strategies (i.e., taking the latest $K$ observations as history, denoted as \texttt{w/o rec}). The learning curves of different strategies are in Figure~\ref{fig:ab}~(right). In general, the impact of history window size is highly game-dependent, but the pure recency based ones do not differ significantly from RC-DQN at the beginning of learning. The issues of pure recency based strategy are: (1) limited additional information about objects provided by successive observations; and (2) higher variance of retrieved observations due to policy changes.

% ----------------------------------------------------

\section{Conclusion}
We formulate the general IF game playing as MPRC tasks, enabling an MPRC-style solution to efficiently address the key IF game challenges on the huge combinatorial action space and the partial observability in a unified framework.  Our approaches achieved significant improvement over the previous state-of-the-art on both game scores and training data efficiency. Our formulation also bridges broader NLU/RC techniques to address other critical challenges in IF games for future work, e.g., common-sense reasoning, novelty-driven exploration, and multi-hop inference.

\section*{Acknowledgments}
We would like to thank Matthew Hausknecht for helpful discussions on the Jericho environments.

\bibliographystyle{acl_natbib}
\bibliography{emnlp2020}

\begin{thebibliography}{33}
\expandafter\ifx\csname natexlab\endcsname\relax\def\natexlab#1{#1}\fi

\bibitem[{Adhikari et~al.(2020)Adhikari, Yuan, C{\^o}t{\'e}, Zelinka, Rondeau,
  Laroche, Poupart, Tang, Trischler, and Hamilton}]{adhikari2020learning}
Ashutosh Adhikari, Xingdi Yuan, Marc-Alexandre C{\^o}t{\'e}, Mikul{\'a}{\v{s}}
  Zelinka, Marc-Antoine Rondeau, Romain Laroche, Pascal Poupart, Jian Tang,
  Adam Trischler, and William~L Hamilton. 2020.
\newblock Learning dynamic knowledge graphs to generalize on text-based games.
\newblock \emph{arXiv preprint arXiv:2002.09127}.

\bibitem[{Ammanabrolu and Hausknecht(2020)}]{ammanabrolu2020graph}
Prithviraj Ammanabrolu and Matthew Hausknecht. 2020.
\newblock Graph constrained reinforcement learning for natural language action
  spaces.
\newblock \emph{arXiv}, pages arXiv--2001.

\bibitem[{Ammanabrolu and Riedl(2019)}]{ammanabrolu2019playing}
Prithviraj Ammanabrolu and Mark Riedl. 2019.
\newblock Playing text-adventure games with graph-based deep reinforcement
  learning.
\newblock In \emph{Proceedings of the 2019 Conference of the North American
  Chapter of the Association for Computational Linguistics: Human Language
  Technologies, Volume 1 (Long and Short Papers)}, pages 3557--3565.

\bibitem[{Asai et~al.(2019)Asai, Hashimoto, Hajishirzi, Socher, and
  Xiong}]{asai2019learning}
Akari Asai, Kazuma Hashimoto, Hannaneh Hajishirzi, Richard Socher, and Caiming
  Xiong. 2019.
\newblock Learning to retrieve reasoning paths over wikipedia graph for
  question answering.
\newblock \emph{arXiv preprint arXiv:1911.10470}.

\bibitem[{Ba et~al.(2016)Ba, Kiros, and Hinton}]{ba2016layer}
Jimmy~Lei Ba, Jamie~Ryan Kiros, and Geoffrey~E Hinton. 2016.
\newblock Layer normalization.
\newblock \emph{arXiv preprint arXiv:1607.06450}.

\bibitem[{Chen et~al.(2017)Chen, Fisch, Weston, and Bordes}]{chen2017reading}
Danqi Chen, Adam Fisch, Jason Weston, and Antoine Bordes. 2017.
\newblock Reading wikipedia to answer open-domain questions.
\newblock In \emph{Proceedings of the 55th Annual Meeting of the Association
  for Computational Linguistics (Volume 1: Long Papers)}, pages 1870--1879.

\bibitem[{Cho et~al.(2014)Cho, Van~Merri{\"e}nboer, Bahdanau, and
  Bengio}]{cho2014properties}
Kyunghyun Cho, Bart Van~Merri{\"e}nboer, Dzmitry Bahdanau, and Yoshua Bengio.
  2014.
\newblock On the properties of neural machine translation: Encoder-decoder
  approaches.
\newblock \emph{arXiv preprint arXiv:1409.1259}.

\bibitem[{C{\^o}t{\'e} et~al.(2018)C{\^o}t{\'e}, K{\'a}d{\'a}r, Yuan, Kybartas,
  Barnes, Fine, Moore, Hausknecht, El~Asri, Adada et~al.}]{cote2018textworld}
Marc-Alexandre C{\^o}t{\'e}, {\'A}kos K{\'a}d{\'a}r, Xingdi Yuan, Ben Kybartas,
  Tavian Barnes, Emery Fine, James Moore, Matthew Hausknecht, Layla El~Asri,
  Mahmoud Adada, et~al. 2018.
\newblock Textworld: A learning environment for text-based games.
\newblock In \emph{Workshop on Computer Games}, pages 41--75. Springer.

\bibitem[{Devlin et~al.(2019)Devlin, Chang, Lee, and
  Toutanova}]{devlin2019bert}
Jacob Devlin, Ming-Wei Chang, Kenton Lee, and Kristina Toutanova. 2019.
\newblock Bert: Pre-training of deep bidirectional transformers for language
  understanding.
\newblock In \emph{Proceedings of the 2019 Conference of the North American
  Chapter of the Association for Computational Linguistics: Human Language
  Technologies, Volume 1 (Long and Short Papers)}, pages 4171--4186.

\bibitem[{Ding et~al.(2019)Ding, Zhou, Chen, Yang, and
  Tang}]{ding2019cognitive}
Ming Ding, Chang Zhou, Qibin Chen, Hongxia Yang, and Jie Tang. 2019.
\newblock Cognitive graph for multi-hop reading comprehension at scale.
\newblock In \emph{Proceedings of ACL 2019}.

\bibitem[{Godbole et~al.(2019)Godbole, Kavarthapu, Das, Gong, Singhal, Zamani,
  Yu, Gao, Guo, Zaheer et~al.}]{godbole2019multi}
Ameya Godbole, Dilip Kavarthapu, Rajarshi Das, Zhiyu Gong, Abhishek Singhal,
  Hamed Zamani, Mo~Yu, Tian Gao, Xiaoxiao Guo, Manzil Zaheer, et~al. 2019.
\newblock Multi-step entity-centric information retrieval for multi-hop
  question answering.
\newblock \emph{arXiv preprint arXiv:1909.07598}.

\bibitem[{Hausknecht et~al.(2019{\natexlab{a}})Hausknecht, Ammanabrolu,
  C{\^o}t{\'e}, and Yuan}]{hausknecht2019interactive}
Matthew Hausknecht, Prithviraj Ammanabrolu, Marc-Alexandre C{\^o}t{\'e}, and
  Xingdi Yuan. 2019{\natexlab{a}}.
\newblock Interactive fiction games: A colossal adventure.
\newblock \emph{arXiv preprint arXiv:1909.05398}.

\bibitem[{Hausknecht et~al.(2019{\natexlab{b}})Hausknecht, Loynd, Yang,
  Swaminathan, and Williams}]{hausknecht2019nail}
Matthew Hausknecht, Ricky Loynd, Greg Yang, Adith Swaminathan, and Jason~D
  Williams. 2019{\natexlab{b}}.
\newblock Nail: A general interactive fiction agent.
\newblock \emph{arXiv preprint arXiv:1902.04259}.

\bibitem[{He et~al.(2016)He, Chen, He, Gao, Li, Deng, and
  Ostendorf}]{he2016deep}
Ji~He, Jianshu Chen, Xiaodong He, Jianfeng Gao, Lihong Li, Li~Deng, and Mari
  Ostendorf. 2016.
\newblock Deep reinforcement learning with a natural language action space.
\newblock In \emph{Proceedings of the 54th Annual Meeting of the Association
  for Computational Linguistics (Volume 1: Long Papers)}, pages 1621--1630.

\bibitem[{Joshi et~al.(2017)Joshi, Choi, Weld, and
  Zettlemoyer}]{joshi2017triviaqa}
Mandar Joshi, Eunsol Choi, Daniel~S Weld, and Luke Zettlemoyer. 2017.
\newblock Triviaqa: A large scale distantly supervised challenge dataset for
  reading comprehension.
\newblock In \emph{Proceedings of the 55th Annual Meeting of the Association
  for Computational Linguistics (Volume 1: Long Papers)}, pages 1601--1611.

\bibitem[{Kwiatkowski et~al.(2019)Kwiatkowski, Palomaki, Redfield, Collins,
  Parikh, Alberti, Epstein, Polosukhin, Devlin, Lee
  et~al.}]{kwiatkowski2019natural}
Tom Kwiatkowski, Jennimaria Palomaki, Olivia Redfield, Michael Collins, Ankur
  Parikh, Chris Alberti, Danielle Epstein, Illia Polosukhin, Jacob Devlin,
  Kenton Lee, et~al. 2019.
\newblock Natural questions: a benchmark for question answering research.
\newblock \emph{Transactions of the Association for Computational Linguistics},
  7:453--466.

\bibitem[{Lin et~al.(2018)Lin, Ji, Liu, and Sun}]{lin2018denoising}
Yankai Lin, Haozhe Ji, Zhiyuan Liu, and Maosong Sun. 2018.
\newblock Denoising distantly supervised open-domain question answering.
\newblock In \emph{Proceedings of the 56th Annual Meeting of the Association
  for Computational Linguistics (Volume 1: Long Papers)}, pages 1736--1745.

\bibitem[{Maas et~al.(2013)Maas, Hannun, and Ng}]{maas2013rectifier}
Andrew~L Maas, Awni~Y Hannun, and Andrew~Y Ng. 2013.
\newblock Rectifier nonlinearities improve neural network acoustic models.
\newblock In \emph{Proc. icml}, volume~30, page~3.

\bibitem[{Min et~al.(2019{\natexlab{a}})Min, Chen, Hajishirzi, and
  Zettlemoyer}]{min2019discrete}
Sewon Min, Danqi Chen, Hannaneh Hajishirzi, and Luke Zettlemoyer.
  2019{\natexlab{a}}.
\newblock A discrete hard em approach for weakly supervised question answering.
\newblock In \emph{Proceedings of the 2019 Conference on Empirical Methods in
  Natural Language Processing and the 9th International Joint Conference on
  Natural Language Processing (EMNLP-IJCNLP)}, pages 2844--2857.

\bibitem[{Min et~al.(2019{\natexlab{b}})Min, Chen, Zettlemoyer, and
  Hajishirzi}]{min2019knowledge}
Sewon Min, Danqi Chen, Luke Zettlemoyer, and Hannaneh Hajishirzi.
  2019{\natexlab{b}}.
\newblock Knowledge guided text retrieval and reading for open domain question
  answering.
\newblock \emph{arXiv preprint arXiv:1911.03868}.

\bibitem[{Mnih et~al.(2015)Mnih, Kavukcuoglu, Silver, Rusu, Veness, Bellemare,
  Graves, Riedmiller, Fidjeland, Ostrovski et~al.}]{mnih2015human}
Volodymyr Mnih, Koray Kavukcuoglu, David Silver, Andrei~A Rusu, Joel Veness,
  Marc~G Bellemare, Alex Graves, Martin Riedmiller, Andreas~K Fidjeland, Georg
  Ostrovski, et~al. 2015.
\newblock Human-level control through deep reinforcement learning.
\newblock \emph{Nature}, 518(7540):529--533.

\bibitem[{Mou et~al.(2020)Mou, Yu, Yao, Yang, Guo, Potdar, and
  Su}]{mou2020frustratingly}
Xiangyang Mou, Mo~Yu, Bingsheng Yao, Chenghao Yang, Xiaoxiao Guo, Saloni
  Potdar, and Hui Su. 2020.
\newblock Frustratingly hard evidence retrieval for qa over books.
\newblock In \emph{Proceedings of the First Joint Workshop on Narrative
  Understanding, Storylines, and Events}, pages 108--113.

\bibitem[{Narasimhan et~al.(2015)Narasimhan, Kulkarni, and
  Barzilay}]{narasimhan2015language}
Karthik Narasimhan, Tejas Kulkarni, and Regina Barzilay. 2015.
\newblock Language understanding for text-based games using deep reinforcement
  learning.
\newblock In \emph{Proceedings of the 2015 Conference on Empirical Methods in
  Natural Language Processing}, pages 1--11.

\bibitem[{Pennington et~al.(2014)Pennington, Socher, and
  Manning}]{pennington2014glove}
Jeffrey Pennington, Richard Socher, and Christopher~D Manning. 2014.
\newblock Glove: Global vectors for word representation.
\newblock In \emph{Proceedings of the 2014 conference on empirical methods in
  natural language processing (EMNLP)}, pages 1532--1543.

\bibitem[{Peters et~al.(2018)Peters, Neumann, Iyyer, Gardner, Clark, Lee, and
  Zettlemoyer}]{peters2018deep}
Matthew~E Peters, Mark Neumann, Mohit Iyyer, Matt Gardner, Christopher Clark,
  Kenton Lee, and Luke Zettlemoyer. 2018.
\newblock Deep contextualized word representations.
\newblock In \emph{Proceedings of NAACL-HLT}, pages 2227--2237.

\bibitem[{Radford et~al.(2019)Radford, Wu, Child, Luan, Amodei, and
  Sutskever}]{radford2019language}
Alec Radford, Jeff Wu, Rewon Child, David Luan, Dario Amodei, and Ilya
  Sutskever. 2019.
\newblock Language models are unsupervised multitask learners.

\bibitem[{Rajpurkar et~al.(2016)Rajpurkar, Zhang, Lopyrev, and
  Liang}]{rajpurkar2016squad}
Pranav Rajpurkar, Jian Zhang, Konstantin Lopyrev, and Percy Liang. 2016.
\newblock Squad: 100,000+ questions for machine comprehension of text.
\newblock In \emph{Proceedings of the 2016 Conference on Empirical Methods in
  Natural Language Processing}, pages 2383--2392.

\bibitem[{Seo et~al.(2016)Seo, Kembhavi, Farhadi, and
  Hajishirzi}]{seo2016bidirectional}
Minjoon Seo, Aniruddha Kembhavi, Ali Farhadi, and Hannaneh Hajishirzi. 2016.
\newblock Bidirectional attention flow for machine comprehension.

\bibitem[{Wang and Jiang(2016)}]{wang2016machine}
Shuohang Wang and Jing Jiang. 2016.
\newblock Machine comprehension using match-lstm and answer pointer.
\newblock \emph{arXiv preprint arXiv:1608.07905}.

\bibitem[{Wang et~al.(2018)Wang, Yu, Guo, Wang, Klinger, Zhang, Chang, Tesauro,
  Zhou, and Jiang}]{wang2018r}
Shuohang Wang, Mo~Yu, Xiaoxiao Guo, Zhiguo Wang, Tim Klinger, Wei Zhang, Shiyu
  Chang, Gerry Tesauro, Bowen Zhou, and Jing Jiang. 2018.
\newblock R 3: Reinforced ranker-reader for open-domain question answering.
\newblock In \emph{Thirty-Second AAAI Conference on Artificial Intelligence}.

\bibitem[{Wang et~al.(2017)Wang, Yu, Jiang, Zhang, Guo, Chang, Wang, Klinger,
  Tesauro, and Campbell}]{wang2017evidence}
Shuohang Wang, Mo~Yu, Jing Jiang, Wei Zhang, Xiaoxiao Guo, Shiyu Chang, Zhiguo
  Wang, Tim Klinger, Gerald Tesauro, and Murray Campbell. 2017.
\newblock Evidence aggregation for answer re-ranking in open-domain question
  answering.
\newblock \emph{arXiv preprint arXiv:1711.05116}.

\bibitem[{Yu et~al.(2018)Yu, Dohan, Luong, Zhao, Chen, Norouzi, and
  Le}]{yu2018qanet}
Adams~Wei Yu, David Dohan, Minh-Thang Luong, Rui Zhao, Kai Chen, Mohammad
  Norouzi, and Quoc~V Le. 2018.
\newblock Qanet: Combining local convolution with global self-attention for
  reading comprehension.
\newblock \emph{arXiv preprint arXiv:1804.09541}.

\bibitem[{Zahavy et~al.(2018)Zahavy, Haroush, Merlis, Mankowitz, and
  Mannor}]{zahavy2018learn}
Tom Zahavy, Matan Haroush, Nadav Merlis, Daniel~J Mankowitz, and Shie Mannor.
  2018.
\newblock Learn what not to learn: Action elimination with deep reinforcement
  learning.
\newblock In \emph{Advances in Neural Information Processing Systems}, pages
  3562--3573.

\end{thebibliography}

\newpage 
\appendix

\onecolumn

\section{Hyper-parameters and Reproducibility Checklist}
\label{app:checklist}

\paragraph{Implementation libraries.} We implemented our models in PyTorch v1.4.0. The Jericho environment tested in our experiments is v2.4.3/2.4.2. The spaCy library version is v2.2.3. 

\paragraph{Computing infrastructure.} The experiments run on servers with Intel(R) Xeon(R) CPU E5-2650 v4 and Nvidia GPUs (can be one of Tesla P100, V100, or K80). The allocated RAM is 150G.

\paragraph{Training time. } The training time is game-specific, ranging from 8 hours to 30 hours. The main factor in the time variance is the size of the combinatorial action space. 

\paragraph{Hyper-parameters. } We did not conduct extensive hyper-parameter tuning. We only tuned the learning rate of Adam from $\big[0.001, 0.0003, 0.0001\big]$ and selected $0.0001$ based on its performance on the game \textit{Zork1}. 

\paragraph{Architecture of Transformer-based block encoder.}
Following QANet~\cite{yu2018qanet}, our Transformer-based block encoder consists of 1) position encoder layer, 2) layer normalization layer, 3) depthwise separable convolution layer, 4) layer normalization, 5) multi-head attention (4-head), 6) layer normalization, and 7) feedforward layer in order. The head number of multi-head attention is reduced from 8 to 4 due to memory constraints. 

\section{Learning Curves}
\label{app:curves}

\begin{figure}[h]
\begin{subfigure}{.32\textwidth}
  \centering
  \includegraphics[width=.95\linewidth]{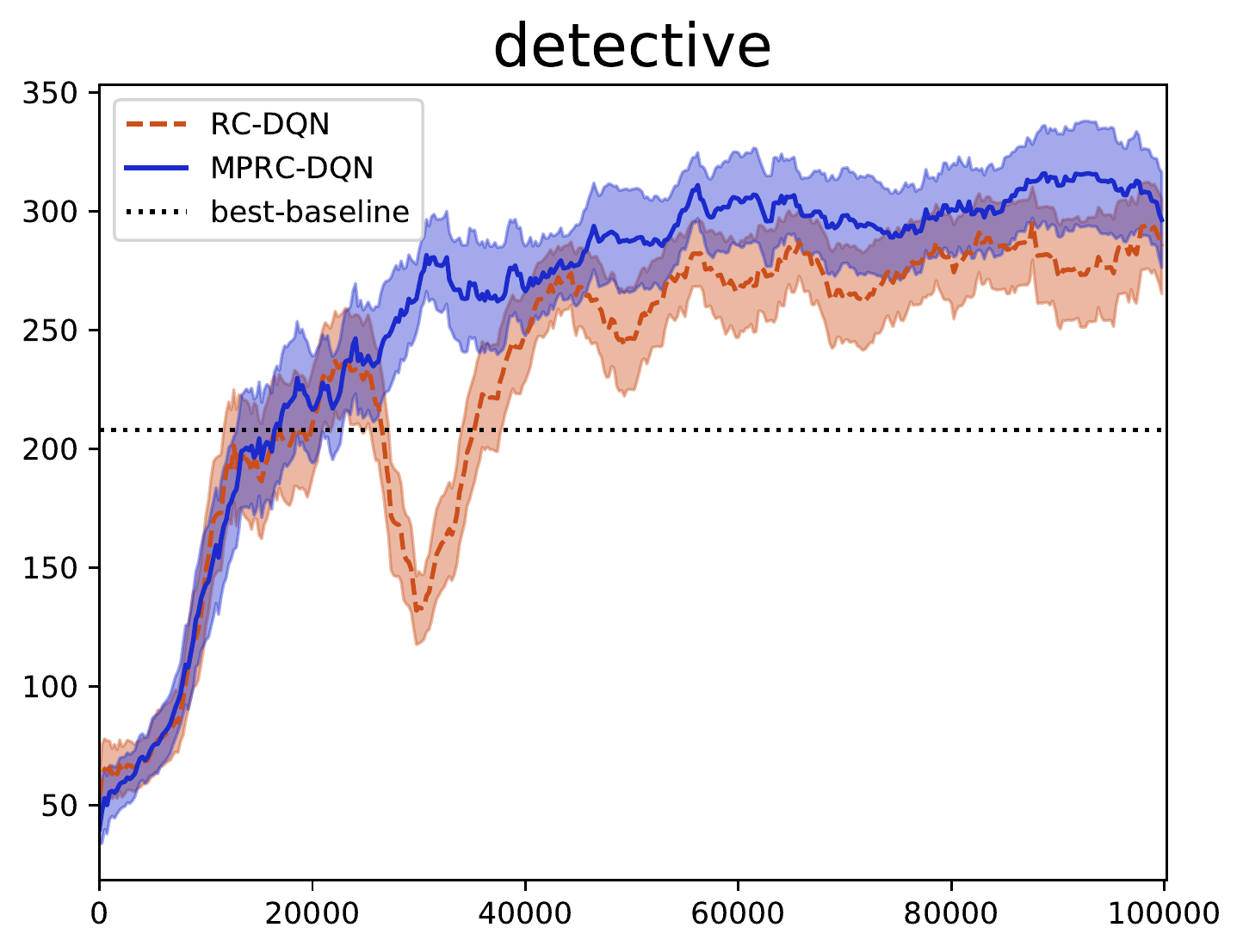}
\end{subfigure}%
\begin{subfigure}{.32\textwidth}
  \centering
  \includegraphics[width=.95\linewidth]{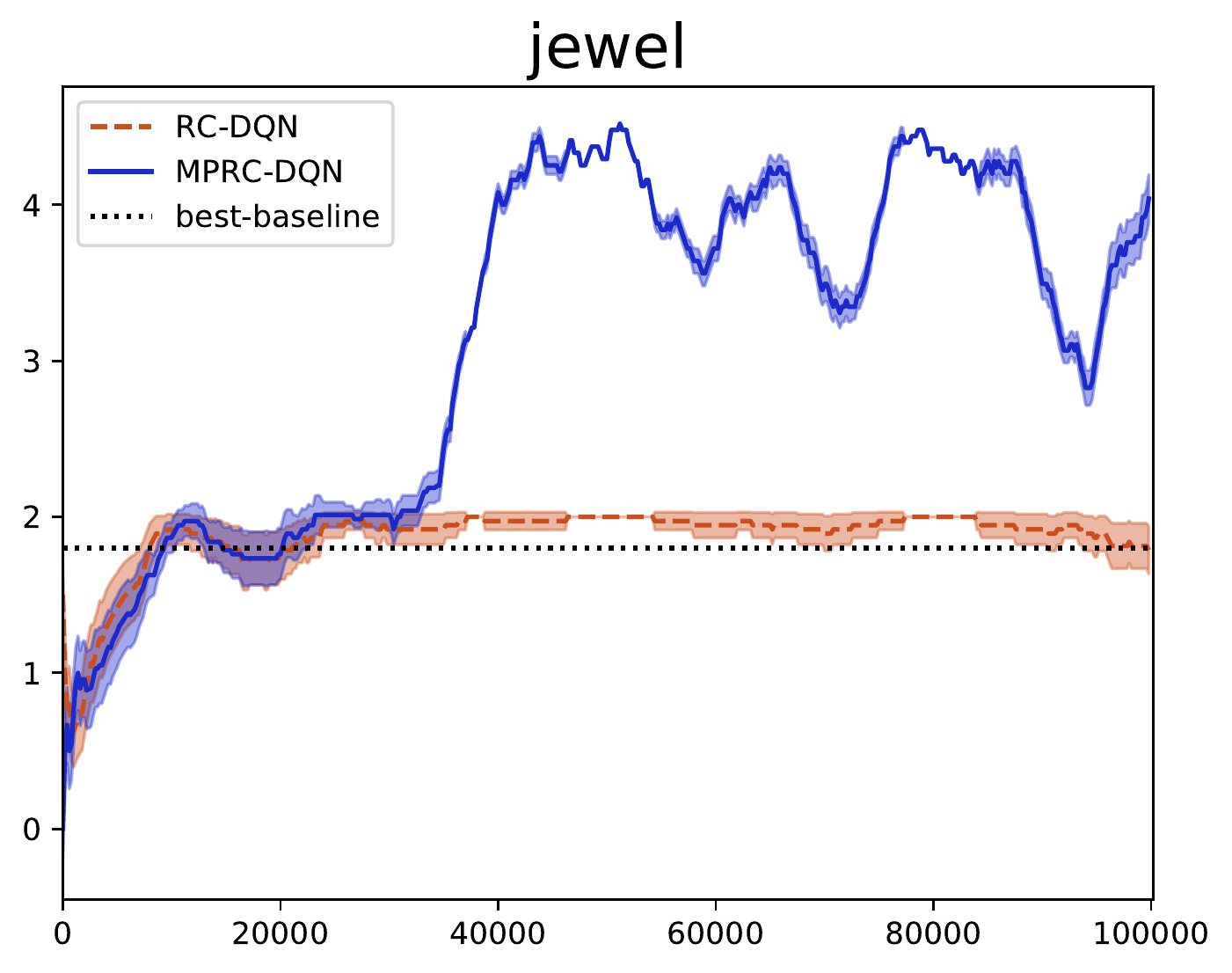}
\end{subfigure}%
\begin{subfigure}{.32\textwidth}
  \centering
  \includegraphics[width=.95\linewidth]{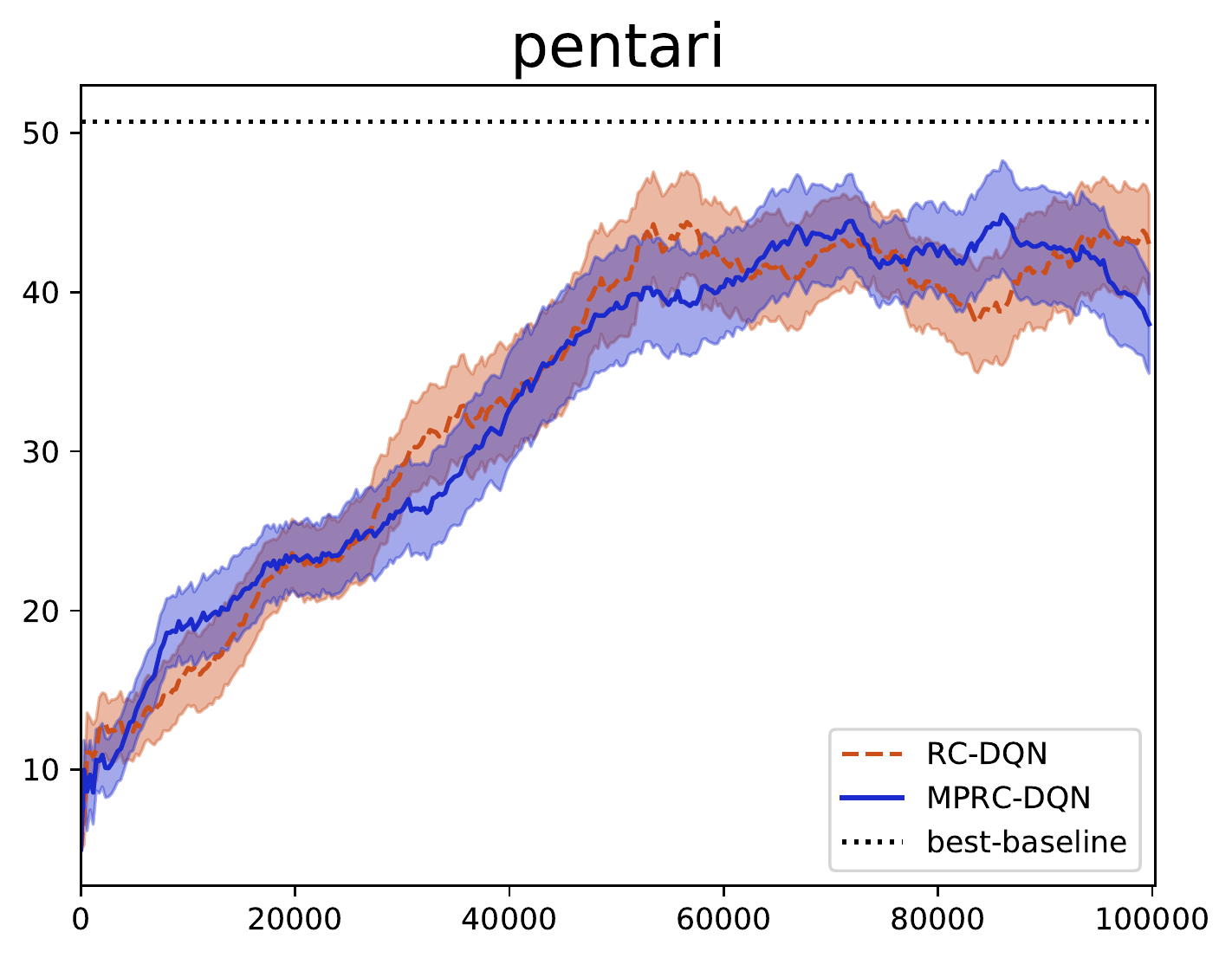}
\end{subfigure}
\begin{subfigure}{.32\textwidth}
  \centering
  \includegraphics[width=.95\linewidth]{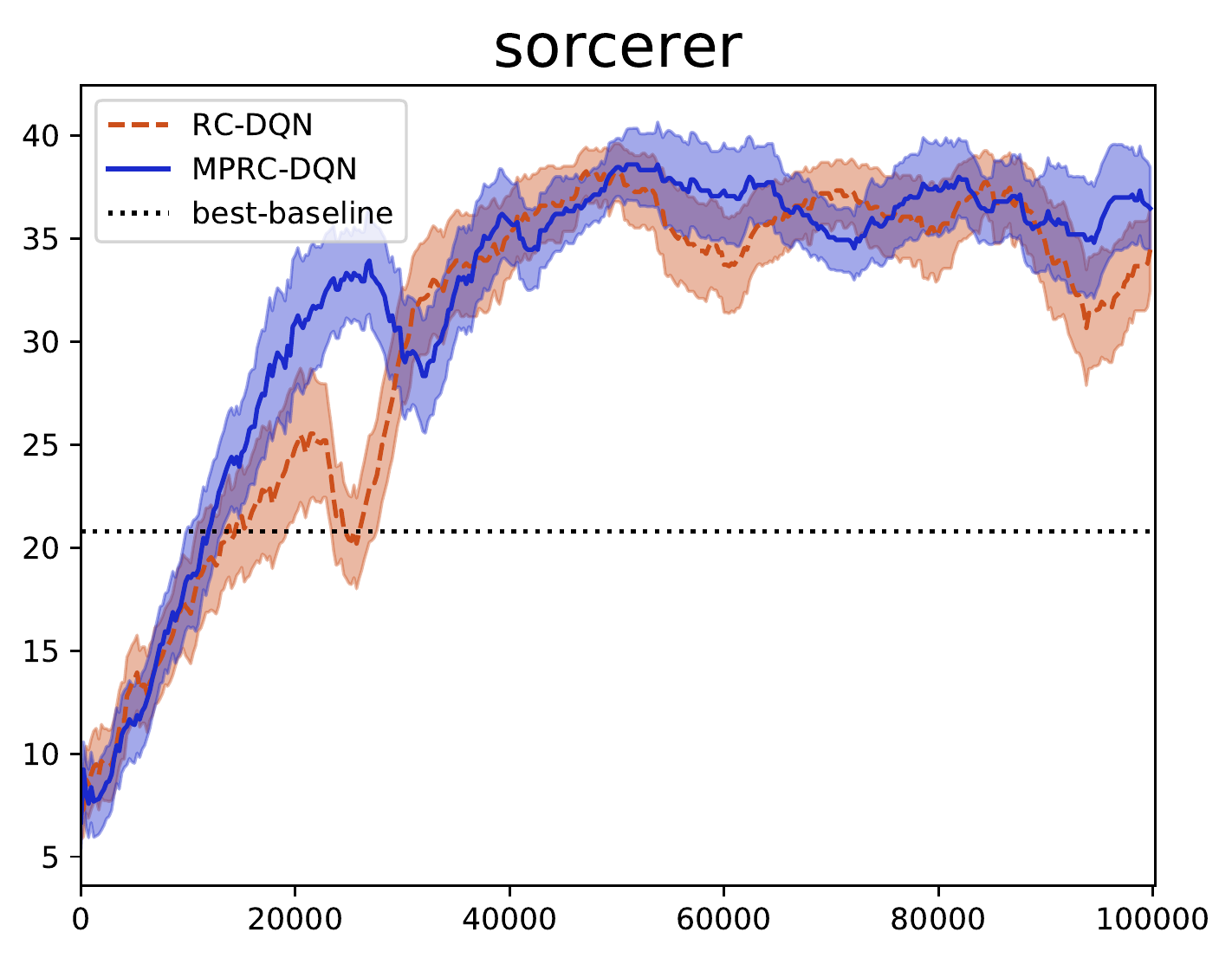}
\end{subfigure}%
\begin{subfigure}{.32\textwidth}
  \centering
  \includegraphics[width=0.95\linewidth]{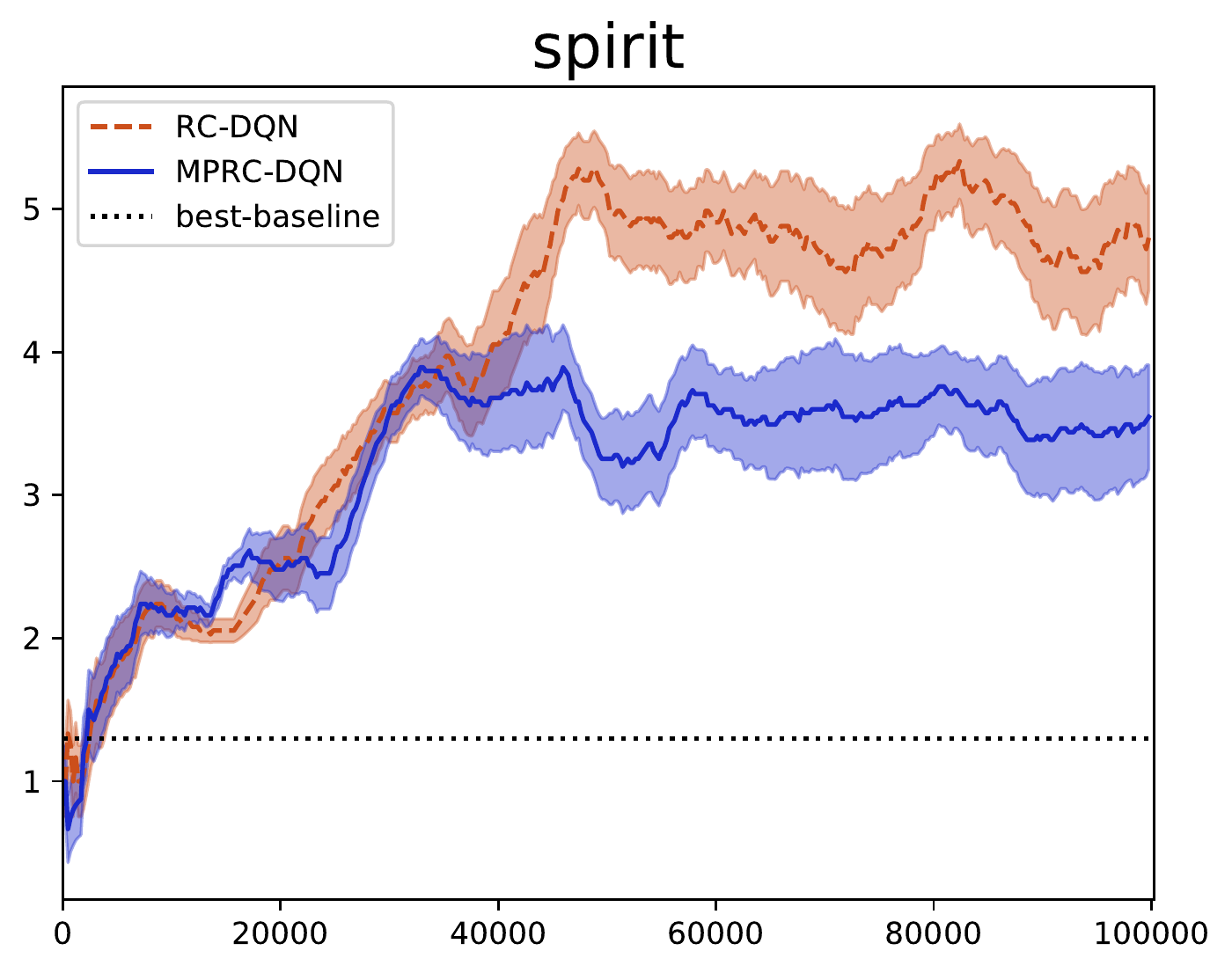}
\end{subfigure}%
\begin{subfigure}{.32\textwidth}
  \centering
  \includegraphics[width=.95\linewidth]{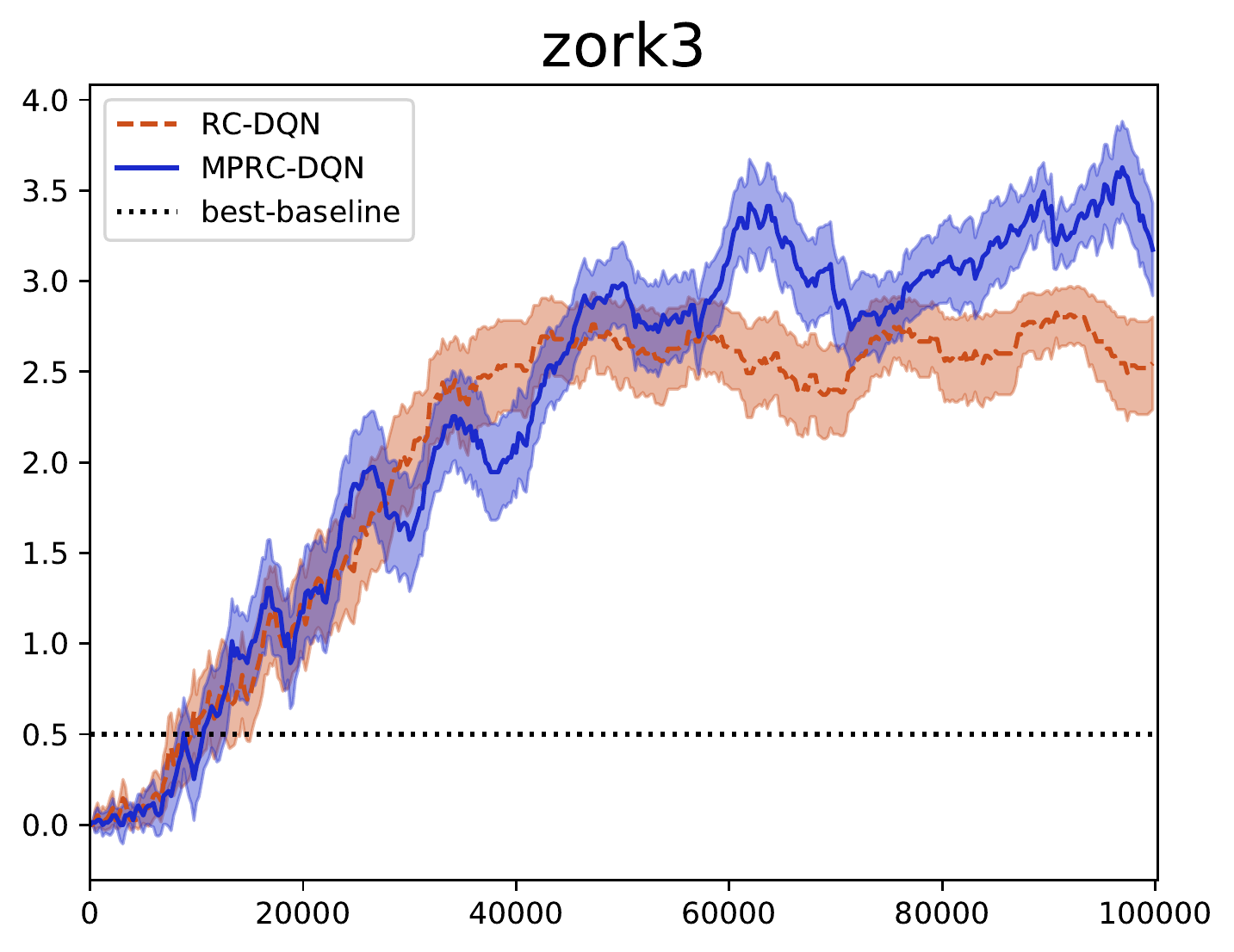}
\end{subfigure}
\caption{\small Learning curves of our MPRC-DQN and RC-DQN agents on various IF games. The best performing baseline results are also included as horizontal lines. }
\label{fig:learning_curves}
\end{figure}

Figure~\ref{fig:learning_curves} shows the learning curves of our agents. The learning curves show that the differences among MPRC-DQN, RC-DQN, and the baselines are statistically significant. The learning curves also indicate that MPRC-DQN and RC-DQN outperform the most competitive baseline on most games at a very early stage of training. 

\section{Game Play Trajectories}
\label{app:traj}

\subsection{Detective}
[step 0] Obs: 
[Type "help" for more information about this version]  Detective By Matt Barringer. Ported by Stuart Moore. Stuart_Moore@my deja.com Release 1 / Serial number 000715 / Inform v6.21 Library 6/10 SD  $\langle\langle$ Chief's office $\rangle\rangle$ You are standing in the Chief's office. He is telling you "The Mayor was murdered yeaterday night at 12:03 am. I want you to solve it before we get any bad publicity or the FBI has to come in. "Yessir!" You reply. He hands you a sheet of paper. Once you have read it, go north or west.  You can see a piece of white paper here.  [Your score has just gone up by ten points.]

[step 1] Act: blow paper

[step 1] Reward:10, Cumulative Rewards:10

[step 2] Obs: 
(first taking the piece of white paper)  [Your score has just gone up by ten points.] You can't usefully blow that.

[step 2] Act: enter west

[step 2] Reward:0, Cumulative Rewards:10

[step 3] Obs: 
$\langle\langle$ Closet $\rangle\rangle$ You are in a closet. There is a gun on the floor. Better get it. To exit, go east.  You can see a small black pistol here.

[step 3] Act: blow black

[step 3] Reward:10, Cumulative Rewards:20

[step 4] Obs: 
(first taking the small black pistol)  [Your score has just gone up by ten points.] You can't usefully blow that.

[step 4] Act: east

[step 4] Reward:0, Cumulative Rewards:20

[step 5] Obs: 
$\langle\langle$ Chief's office $\rangle\rangle$ You are standing in the Chief's office. He is telling you "The Mayor was murdered yeaterday night at 12:03 am. I want you to solve it before we get any bad publicity or the FBI has to come in. "Yessir!" You reply. He hands you a sheet of paper. Once you have read it, go north or west.

[step 5] Act: north

[step 5] Reward:0, Cumulative Rewards:20

[step 6] Obs: 
$\langle\langle$ Outside $\rangle\rangle$ You are outside in the cold. To the east is a dead end. To the west is the rest of the street. Papers are blowing around. It's amazingly cold for this time of year.

[step 6] Act: enter west

[step 6] Reward:10, Cumulative Rewards:30

[step 7] Obs: 
$\langle\langle$ Outside $\rangle\rangle$ You are still on the streets. To the north is a restraunt where the mayor ate often. To the east is the Mayor's home.  [Your score has just gone up by ten points.]

[step 7] Act: east

[step 7] Reward:10, Cumulative Rewards:40

[step 8] Obs: 
$\langle\langle$ Mayor's house $\rangle\rangle$ You are in the house, at the scene of the crime. You enter and flash your badge before a cop. He admits you. To the north is the upstairs. To the east is the living room and to the west is the dining room.  [Your score has just gone up by ten points.]

[step 8] Act: enter north

[step 8] Reward:10, Cumulative Rewards:50

[step 9] Obs: 
$\langle\langle$ Upstairs hallway $\rangle\rangle$ You are in the hallway of the large house of the mayor. It is an amazingly large house. You can go north, south, east or west.  [Your score has just gone up by ten points.]

[step 9] Act: north

[step 9] Reward:0, Cumulative Rewards:50

[step 10] Obs: 
$\langle\langle$ Hallway $\rangle\rangle$ You are at the end of the hallway. To the north is a room, while to the west is the rest of the hallway.

[step 10] Act: enter west

[step 10] Reward:0, Cumulative Rewards:50

[step 11] Obs: 
$\langle\langle$ Hallway $\rangle\rangle$ You are in the hallway. To the north is more hallway, and to the east is a door marked "Guests".

[step 11] Act: enter north

[step 11] Reward:0, Cumulative Rewards:50

[step 12] Obs: 
$\langle\langle$ Hallway $\rangle\rangle$ You are STILL in the hallway. There is EVEN MORE hallway to the north, and a room to the west and a room to the east of you.

[step 12] Act: enter north

[step 12] Reward:0, Cumulative Rewards:50

[step 13] Obs: 
$\langle\langle$ Hallway $\rangle\rangle$ You are still in the hallway. You can go north to where there is a police officer who will let you outside, or you can go east or west.

[step 13] Act: north

[step 13] Reward:10, Cumulative Rewards:60

[step 14] Obs: 
$\langle\langle$ Outside $\rangle\rangle$ You pass the guard. He nods at you. You are now outside standing on the street. You can go north and east, your choice. To the north is more of the street, and to the east is a video store.  [Your score has just gone up by ten points.]

[step 14] Act: east

[step 14] Reward:0, Cumulative Rewards:60

[step 15] Obs: 
$\langle\langle$ Video Store $\rangle\rangle$ You are in a video store called Brickbuster Video. There are about 3,000 videos here. You can go north, or east.

[step 15] Act: drop paper

[step 15] Reward:0, Cumulative Rewards:60

[step 16] Obs: 
Dropped.

[step 16] Act: east

[step 16] Reward:10, Cumulative Rewards:70

[step 17] Obs: 
$\langle\langle$ Video Store $\rangle\rangle$ You are still in the video store. You can go north, or east.  [Your score has just gone up by ten points.]

[step 17] Act: east

[step 17] Reward:0, Cumulative Rewards:70

[step 18] Obs: 
$\langle\langle$ Outside $\rangle\rangle$ You are outside. You can go north, south, east or west.

[step 18] Act: enter south

[step 18] Reward:0, Cumulative Rewards:70

[step 19] Obs: 
$\langle\langle$ McDonalds $\rangle\rangle$ You are in a McDonalds. You pay the guy behind the counter. Now there is a hamburger there. When you have picked it up, go north.  You can see a food hamburger here.

[step 19] Act: drop black

[step 19] Reward:0, Cumulative Rewards:70

[step 20] Obs: 
Dropped.

[step 20] Act: blow black

[step 20] Reward:0, Cumulative Rewards:70

[step 21] Obs: 
(first taking the small black pistol) You can't usefully blow that.

[step 21] Act: blow food

[step 21] Reward:10, Cumulative Rewards:80

[step 22] Obs: 
(first taking the food hamburger)  [Your score has just gone up by ten points.] You can't usefully blow that.

[step 22] Act: north

[step 22] Reward:0, Cumulative Rewards:80

[step 23] Obs: 
$\langle\langle$ Outside $\rangle\rangle$ You are outside. You can go north, south, east or west.

[step 23] Act: north

[step 23] Reward:10, Cumulative Rewards:90

[step 24] Obs: 
$\langle\langle$ Outside $\rangle\rangle$ You are still outside. You hit a dead end, then notice that you can go east only.  [Your score has just gone up by ten points.]

[step 24] Act: east

[step 24] Reward:10, Cumulative Rewards:100

[step 25] Obs: 
$\langle\langle$ Music Store $\rangle\rangle$ You are in a music store. You ask the man behind the counter if he knew any information. "Uhh... nope! But the guy back there might be able ta help." You politly thank him and head to the back. You can only go north.  [Your score has just gone up by ten points.]

[step 25] Act: north

[step 25] Reward:10, Cumulative Rewards:110

[step 26] Obs: 
$\langle\langle$ Back of Music Store $\rangle\rangle$ You are in the back of the music store. You ask the guy who's looking at the cool tapes. He looks up at you. "Duh.. no... don't t'ink so... lemme see..." You decide that he's no help. To the west there is a dazed looking man and to the north there is an exit.  [Your score has just gone up by ten points.]

[step 26] Act: enter north

[step 26] Reward:10, Cumulative Rewards:120

[step 27] Obs: 
$\langle\langle$ Alley $\rangle\rangle$ You are in an alley. A drunken man stagers up to you and says "Boycott FDR! $\langle$ HICKUP $\rangle$.". You just walk away. You can go north, east or west. Your call.  [Your score has just gone up by ten points.]

[step 27] Act: north

[step 27] Reward:10, Cumulative Rewards:130

[step 28] Obs: 
$\langle\langle$ Police Station $\rangle\rangle$ You are in the 3rd precinct police station. This isn't your station. You get admitance from the guy at the desk and go to the holding cells. You ask each offender if they know anything. You promise a lighter sentence for the ones who help. But one guy really sets you straight. "I got caught wit' t'ree ounces o' crack. I'm supposed to get 20 years but I'll be out in 2. You can't make me talk cuz it don't matter to me. If I squeal, da guys who did it are gonna come lookin' for me. I know but I ain't gonna tell ya. Now git outta my face.". You are surprised but used to it. You can go north to the outside, south to go back to the alley and west or east to talk to more guys.  [Your score has just gone up by ten points.]

[step 28] Act: north

[step 28] Reward:10, Cumulative Rewards:140

[step 29] Obs: 
$\langle\langle$ Outside $\rangle\rangle$ You are outside. it's bitter cold and you pull your jacket around yourself. To the north is a nice, warm Holiday Inn hotel, where the killer is rumoured to be staying. Or you could go to his favourite hang out, the Wall, to the west, or to the east is the place where he is supposed to be working, the Doughnut King.  [Your score has just gone up by ten points.]

[step 29] Act: east

[step 29] Reward:10, Cumulative Rewards:150

[step 30] Obs: 
$\langle\langle$ Doughnut King $\rangle\rangle$ You are in the Doughnut King, where the greasiest doughnuts on earth reside. He isn't here, no one seems to be for that matter, so you should go west.  [Your score has just gone up by ten points.]

[step 30] Act: enter west

[step 30] Reward:0, Cumulative Rewards:150

[step 31] Obs: 
$\langle\langle$ Outside $\rangle\rangle$ You are outside. it's bitter cold and you pull your jacket around yourself. To the north is a nice, warm Holiday Inn hotel, where the killer is rumoured to be staying. Or you could go to his favourite hang out, the Wall, to the west, or to the east is the place where he is supposed to be working, the Doughnut King.

[step 31] Act: enter north

[step 31] Reward:10, Cumulative Rewards:160

[step 32] Obs: 
$\langle\langle$ Holiday Inn $\rangle\rangle$ You are in the Holiday Inn registration room. You talk to some suspicious guys, but they don't talk until you hold your gun to their side. "Alright! Alright! I'll talk! He's on the 15th floor! That's all I can tell ya!" You shove them away. You walk up to the registration desk and show the woman there your badge. She gives you the master ring. You now have access to all of the facilitys on the 15th floor. But the problem is that the 15th floor is the suite level, and there are 30 suites, and 5 pools, 2 saunas and 5 game rooms. Big problem! Well, you have all night. You get a picture of all on the 15th floor, the people up there have to show their drivers license to be admitted, and the license is secretly xeroxed. You look at them all. Well, better get started. You see one person who stands out. You get his room number from the lady. Room 30. Now you have to find it. To get started, go north.  [Your score has just gone up by ten points.]

[step 32] Act: north

[step 32] Reward:10, Cumulative Rewards:170

[step 33] Obs: 
$\langle\langle$ Holiday Inn 15th Floor $\rangle\rangle$ You go up the elevator. When you step out, you see the wallpaper is pink, with little flowers on it. You can go east or west.  [Your score has just gone up by ten points.]

[step 33] Act: enter west

[step 33] Reward:10, Cumulative Rewards:180

[step 34] Obs: 
$\langle\langle$ Hallway $\rangle\rangle$ You are in the hallway. You see many doors... 1... 2... 3... 4... 5... 6... 7... boy, you have a long way to go. You can only go north.  [Your score has just gone up by ten points.]

[step 34] Act: north

[step 34] Reward:10, Cumulative Rewards:190

[step 35] Obs: 
$\langle\langle$ Hallway $\rangle\rangle$ You are still in the hallways. You can go west or east.  [Your score has just gone up by ten points.]

[step 35] Act: enter west

[step 35] Reward:10, Cumulative Rewards:200

[step 36] Obs: 
$\langle\langle$ Hallway $\rangle\rangle$ You are in the hallway. You see numbers flash by as you run through the halls. 19... 20... 21... 22... you are getting close! You can only go north.  [Your score has just gone up by ten points.]

[step 36] Act: north

[step 36] Reward:10, Cumulative Rewards:210

[step 37] Obs: 
$\langle\langle$ Hallway $\rangle\rangle$ You are in the hallway. You feel the heat from the sauna to the west, and to the east is a door marked "Pool A". To the north is more hall.  [Your score has just gone up by ten points.]

[step 37] Act: enter north

[step 37] Reward:100, Cumulative Rewards:310

[step 38] Obs: 
$\langle\langle$ Room 30 $\rangle\rangle$ You enter room 30... after a harrowing gun battle you conk him on the head and take him in. You get promoted and suddenly, with the "Jurrasic Park" theme song playing in your head, you feel proud to be an American. For special info about Exile Games, and to leave this darned game, go up.  [Your score has just gone up by one hundred points.]

[step 38] Act: drop food

[step 38] Reward:0, Cumulative Rewards:310

[step 39] Obs: 
Dropped.

[step 39] Act: drop black

[step 39] Reward:0, Cumulative Rewards:310

[step 40] Obs: 
Dropped.

[step 40] Act: down

[step 40] Reward:0, Cumulative Rewards:310

[step 41] Obs: 
You can't go down from here!  $\langle\langle$ Room 30 $\rangle\rangle$ You enter room 30... after a harrowing gun battle you conk him on the head and take him in. You get promoted and suddenly, with the "Jurrasic Park" theme song playing in your head, you feel proud to be an American. For special info about Exile Games, and to leave this darned game, go up.  You can see a small black pistol and a food hamburger here.

[step 41] Act: down

[step 41] Reward:0, Cumulative Rewards:310

[step 42] Obs: 
You can't go down from here!  $\langle\langle$ Room 30 $\rangle\rangle$ You enter room 30... after a harrowing gun battle you conk him on the head and take him in. You get promoted and suddenly, with the "Jurrasic Park" theme song playing in your head, you feel proud to be an American. For special info about Exile Games, and to leave this darned game, go up.  You can see a small black pistol and a food hamburger here.

[step 42] Act: down

[step 42] Reward:0, Cumulative Rewards:310

[step 43] Obs: 
You can't go down from here!  $\langle\langle$ Room 30 $\rangle\rangle$ You enter room 30... after a harrowing gun battle you conk him on the head and take him in. You get promoted and suddenly, with the "Jurrasic Park" theme song playing in your head, you feel proud to be an American. For special info about Exile Games, and to leave this darned game, go up.  You can see a small black pistol and a food hamburger here.

[step 43] Act: down

[step 43] Reward:0, Cumulative Rewards:310

\texttt{repeatedly trying to go down with some occasional other actions until the max step number is reached.}

\subsection{Dragon}
[step 0] Obs: 
The faces of the council members look grim. Gilgern continues to speak.  "Of course, something must be done soon," he says in that gruff, hearty voice you have come to dislike so much. "Must restore public confidence and encourage the return of people to the land. Can't just leave the place to the likes of dragons and trolls. We must all make money again. Isn't as though it's just arrived, dragon's been there for years. Just that people found out about it, that's all."  Marzipam looks round nervously at the others in the so called Council of the Wise.  "Of course, we can't afford to pay you much. We are just poor men ourselves. Think of this more as a civic duty..." he wheezes.  Around you the other travellers and adventurers shift nervously.  Gilgern hurriedly adjusts his glasses and glances down at the scroll in front of him. "Says here that the Great Worm can be beaten," he goes on. "Sword of Erondil is the only weapon which can penetrate a dragon's scaly hide. Enchanted weapon, you see. Last seen many years ago in the Castle of Abercorn. Big, ugly place. Near the sea, I think, if memory serves me right. Dragon must be getting a bit careless and lazy by now. Been there all that time. Stands to reason. Might get in a lucky hit or something."  Marzipam breaks in. "Or there's the spell, I suppose. The dragon is supposed to be held in place under an old enchantment, though that's perhaps just an old tale. Break the spell and the dragon will leave, the story goes..."  "Well, which of you is interested?" Gilgern looks across at you and the other adventurers. You begin to wish you had never set foot in the town   it just happened to be on your journey, that's all. You've no wish to tangle with dangerous dragons or anything else.  As Gilgern looks down, you suddenly find that the other adventurers have all quietly taken a couple of steps back, leaving you out in the front.  "Splendid", chortles Gilgern before you can say anything, "We've one brave soul willing to risk their all for our little cause. Welcome, stranger..."  Before you know it, you have been whisked away from the town square in a donkey and cart and deposited beside a track.  "Good luck   you'll need it!" calls the driver over his shoulder, as he hurries away.   Welcome to Dragon Adventure. You start your quest on a mountain path in the North East...  (Type INFO at any time for more playing information.)  Dragon Adventure A simple text adventure for players aged 9 and over, by William Stott Release 2 / Serial number 040211 / Inform v6.21 Library 6/10   Rocky Mountains It is a clear, bright morning and all seems peaceful. A tiny sparrowhawk wheels and soars lazily in the cool empty air above. There is a steep pathway to the south leading down into dense woodland, and you can make out an old building by the path. In the distance you can just glimpse the steel grey of the sea, far away to the west.   You can go north, south, east or west.

[step 1] Act: yes

[step 1] Reward:0, Cumulative Rewards:0

[step 2] Obs: 
You run wildly, as fast as you can...  Swampy Ground Water seeps around the soles of your boots and fills your recent footprints. You can hear the croaking of frogs and the occasional bubble of marsh gas coming to the surface. There is an abandoned inn to the east, and a path leads south into the forest. To the west there seems to be a little country road.   You can go south (along a forest path), east (towards an inn) or west (towards a road).  A little man comes up to you and asks if you want to buy a box. You can see a little man and a large cardboard box here.

[step 2] Act: push box to cardboard

[step 2] Reward:0, Cumulative Rewards:0

[step 3] Obs: 
You can't put something inside itself.

[step 3] Act: east

[step 3] Reward:0, Cumulative Rewards:0

[step 4] Obs: 
Abandoned Inn An old, thatched building stands near the path. It seems to be an abandoned inn or tavern, as a faded sign reads 'Troll's Arms'. The walls are a dirty white and the paintwork is starting to peel and blister. The windows are covered in cobwebs and there is the faint odour of mildew and decay. Weeds grow between the flagstones of the path leading up to the door, which leans open. No one seems to have lived here for years. A path leads north into the mountains and there seems to be some swampy ground to the west.   You can go north (to the mountains), west (towards a swamp) or east (into the inn).

[step 4] Act: buy box

[step 4] Reward:1, Cumulative Rewards:1

[step 5] Obs: 
The little man takes the coin and leaves you the cardboard box.  He runs off, laughing...  [Your score has just gone up by one point.]

[step 5] Act: east

[step 5] Reward:0, Cumulative Rewards:1

[step 6] Obs: 
Inside the Inn There are a few patches of mould on the walls and the floorboards creak slightly, but the building seems safe enough. As you move you disturb a thin layer of dust. There are some stone steps leading down, and an open doorway to the outside to the west. You see some tables and chairs. There is a dark, intricately carved wooden chair in the corner behind you.   You can go west (back outside) or down some steps.  You can see a magnificently carved chair here.

[step 6] Act: push chair to box

[step 6] Reward:0, Cumulative Rewards:1

[step 7] Obs: 
That can't contain things.

[step 7] Act: enter west

[step 7] Reward:0, Cumulative Rewards:1

[step 8] Obs: 
Abandoned Inn An old, thatched building stands near the path. It seems to be an abandoned inn or tavern, as a faded sign reads 'Troll's Arms'. The walls are a dirty white and the paintwork is starting to peel and blister. The windows are covered in cobwebs and there is the faint odour of mildew and decay. Weeds grow between the flagstones of the path leading up to the door, which leans open. No one seems to have lived here for years. A path leads north into the mountains and there seems to be some swampy ground to the west.   You can go north (to the mountains), west (towards a swamp) or east (into the inn).

[step 8] Act: drop box

[step 8] Reward:0, Cumulative Rewards:1

[step 9] Obs: 
Dropped.

[step 9] Act: blow box

[step 9] Reward:0, Cumulative Rewards:1

[step 10] Obs: 
(first taking the large cardboard box) You can't usefully blow that.

[step 10] Act: enter west

[step 10] Reward:0, Cumulative Rewards:1

[step 11] Obs: 
Swampy Ground Water seeps around the soles of your boots and fills your recent footprints. You can hear the croaking of frogs and the occasional bubble of marsh gas coming to the surface. There is an abandoned inn to the east, and a path leads south into the forest. To the west there seems to be a little country road.   You can go south (along a forest path), east (towards an inn) or west (towards a road).

[step 11] Act: enter west

[step 11] Reward:0, Cumulative Rewards:1

[step 12] Obs: 
North End of a Road A country road stretches to the south, its gravelled surface crunching slightly beneath your boots. It is bordered by trees and old hedgerows. Towards the east is a patch of swampy ground.   You can go east (towards the swamp) or south (along the road).  You can see a wooden signpost here.

[step 12] Act: enter south

[step 12] Reward:0, Cumulative Rewards:1

[step 13] Obs: 
South End of a Road A gravelled country road leads north. A white painted wooden signpost points west, where you can make out the sea. You can just glimpse the glint of more water through the trees, far to the south.   You can go north (along the road), south (into the trees) or west (towards the sea).  You can see a signpost here.

[step 13] Act: enter west

[step 13] Reward:0, Cumulative Rewards:1

[step 14] Obs: 
Wasteland by a Castle A stretch of uneven waste ground stands closeby to a castle. There are old dunes covered in spiky patches of marram and couch grass. You can make out the old castle and its buildings away to the north. The blue grey of the sea curves around in a wide arc to a beach towards the south.   You can go north (towards the castle), south (to the beach) or east (towards the trees).  You can see an old delivery bicycle here.

[step 14] Act: enter north

[step 14] Reward:0, Cumulative Rewards:1

[step 15] Obs: 
Castle Forge The ruined forge stands just outside the castle. There is no roof, and the remains of thick stone walls are open to the sky. The castle itself is closeby to the north, its stone towers soaring upwards. There is desolate wasteland stretching away to the south.    You can go north (towards the castle itself), or south (to some wasteland).  You can see a brass lamp here.

[step 15] Act: enter north

[step 15] Reward:0, Cumulative Rewards:1

[step 16] Obs: 
Forbidding Castle The hulk of a massive, stone castle towers into the sky above you. The walls are covered in grey lichen and moss, and seem to have stood here since the beginning of time itself. Set into the wall to the north you can see a large oak door, dark and weathered. There is a keyhole surrounded by a black iron plate. The ruins of the castle forge lie to the south. The door is locked, and there is no other way inside.   The only way you can go is south (to the forge).  You can see a small silver bell here.

[step 16] Act: blow bell

[step 16] Reward:0, Cumulative Rewards:1

[step 17] Obs: 
(first taking the small silver bell) You can't usefully blow that.

[step 17] Act: enter south

[step 17] Reward:0, Cumulative Rewards:1

[step 18] Obs: 
Castle Forge The ruined forge stands just outside the castle. There is no roof, and the remains of thick stone walls are open to the sky. The castle itself is closeby to the north, its stone towers soaring upwards. There is desolate wasteland stretching away to the south.    You can go north (towards the castle itself), or south (to some wasteland).  You can see a brass lamp here.

[step 18] Act: ring bell

[step 18] Reward:0, Cumulative Rewards:1

[step 19] Obs: 
You ring the bell...  You hear the note of a horn answering in the distance...

[step 19] Act: south

[step 19] Reward:0, Cumulative Rewards:1

[step 20] Obs: 
Wasteland by a Castle A stretch of uneven waste ground stands closeby to a castle. There are old dunes covered in spiky patches of marram and couch grass. You can make out the old castle and its buildings away to the north. The blue grey of the sea curves around in a wide arc to a beach towards the south.   You can go north (towards the castle), south (to the beach) or east (towards the trees).  You can see an old delivery bicycle here.

[step 20] Act: south

[step 20] Reward:0, Cumulative Rewards:1

[step 21] Obs: 
Pebbled Beach You hear the crashing sound of waves in your ears. White gulls wheel and play in the sky above, and you can taste the salt spray on your lips. Pebbles crunch beneath your feet as you move. You can see the ruins of an old stone lighthouse to the south. There is a wasteland of sand dunes to the north, and forest trees to the east.   You can go north (towards wasteland), south (towards a lighthouse) or east (towards trees).  You can see a parachute here.

[step 21] Act: enter east

[step 21] Reward:0, Cumulative Rewards:1

[step 22] Obs: 
Lake in the Forest Dragonflies hum and hover, and you hear the occasional splash of feeding fish. You have a feeling of tranquility instilled by the placid, lazy waters. There is a little path winding between dense trees towards the south. You can see the start of a deserted gravel road to the north, and you can just make out the sea to the west.   You can go south (further into the trees), north (towards the road) or west (towards the sea).  You can see some long reeds here.

[step 22] Act: blow bell

[step 22] Reward:0, Cumulative Rewards:1

[step 23] Obs: 
You can't usefully blow that.

[step 23] Act: give bell silver

[step 23] Reward:5, Cumulative Rewards:6

[step 24] Obs: 
(to the huge Troll with a club) The Troll stops in surprise and looks at you curiously.  "You're not another of these sneaky thieves, then...?" he says.  "No one has bothered to speak to me for years. Nowadays, people just come to steal and kill.  Thank you for finding my bell. My old hunting horn may be useful to you, I'll let you borrow it, if you like. I'll be off, then."  The Troll leaves you his huge hunting horn, then walks off.  [Your score has just gone up by five points.]

[step 24] Act: enter south

[step 24] Reward:0, Cumulative Rewards:6

[step 25] Obs: 
Tree Stump The twisting path leads through an ancient forest. Red and blue butterflies dance above the grass and you can hear the soft calls of wood pigeons in the trees. The leaves rustle gently in the morning breeze. The pathway leads east deeper into the forest and continues north, winding into the trees. There is an old tree stump next to the path where once a great oak tree must have stood. It seems to be hollow in the centre.   You can go north (along a path) or east (further into the forest).  You can see a hollow tree stump here.

[step 25] Act: examine hollow

[step 25] Reward:1, Cumulative Rewards:7

[step 26] Obs: 
You look carefully at the hollow tree stump.  You have found a box of matches...  [Your score has just gone up by one point.]

[step 26] Act: enter north

[step 26] Reward:0, Cumulative Rewards:7

[step 27] Obs: 
Lake in the Forest Dragonflies hum and hover, and you hear the occasional splash of feeding fish. You have a feeling of tranquility instilled by the placid, lazy waters. There is a little path winding between dense trees towards the south. You can see the start of a deserted gravel road to the north, and you can just make out the sea to the west.   You can go south (further into the trees), north (towards the road) or west (towards the sea).  You can see a big hunting horn and some long reeds here.

[step 27] Act: enter north

[step 27] Reward:0, Cumulative Rewards:7

[step 28] Obs: 
South End of a Road A gravelled country road leads north. A white painted wooden signpost points west, where you can make out the sea. You can just glimpse the glint of more water through the trees, far to the south.   You can go north (along the road), south (into the trees) or west (towards the sea).  You can see a signpost here.

[step 28] Act: enter west

[step 28] Reward:0, Cumulative Rewards:7

[step 29] Obs: 
Wasteland by a Castle A stretch of uneven waste ground stands closeby to a castle. There are old dunes covered in spiky patches of marram and couch grass. You can make out the old castle and its buildings away to the north. The blue grey of the sea curves around in a wide arc to a beach towards the south.   You can go north (towards the castle), south (to the beach) or east (towards the trees).  You can see an old delivery bicycle here.

[step 29] Act: enter north

[step 29] Reward:0, Cumulative Rewards:7

[step 30] Obs: 
Castle Forge The ruined forge stands just outside the castle. There is no roof, and the remains of thick stone walls are open to the sky. The castle itself is closeby to the north, its stone towers soaring upwards. There is desolate wasteland stretching away to the south.    You can go north (towards the castle itself), or south (to some wasteland).  You can see a brass lamp here.

[step 30] Act: enter north

[step 30] Reward:0, Cumulative Rewards:7

[step 31] Obs: 
Forbidding Castle The hulk of a massive, stone castle towers into the sky above you. The walls are covered in grey lichen and moss, and seem to have stood here since the beginning of time itself. Set into the wall to the north you can see a large oak door, dark and weathered. There is a keyhole surrounded by a black iron plate. The ruins of the castle forge lie to the south. The door is locked, and there is no other way inside.   The only way you can go is south (to the forge).

[step 31] Act: enter south

continue with repeating 'enter north' and 'enter south' until time limit 100.

[step 100] Reward:0, Cumulative Rewards:7

[step 101] Obs: 
Lake in the Forest Dragonflies hum and hover, and you hear the occasional splash of feeding fish. You have a feeling of tranquility instilled by the placid, lazy waters. There is a little path winding between dense trees towards the south. You can see the start of a deserted gravel road to the north, and you can just make out the sea to the west.   You can go south (further into the trees), north (towards the road) or west (towards the sea).  You can see a big hunting horn and some long reeds here.

\subsection{Zork1}

[step 0] Obs:
Copyright (c) 1981, 1982, 1983 Infocom, Inc. All rights reserved. ZORK is a registered trademark of Infocom, Inc. Revision 88 / Serial number 840726  West of House You are standing in an open field west of a white house, with a boarded front door. There is a small mailbox here.

[step 1] Act: west

[step 1] Reward:0, Cumulative Rewards:0

[step 2] Obs:
Forest This is a forest, with trees in all directions. To the east, there appears to be sunlight.

[step 2] Act: east

[step 2] Reward:0, Cumulative Rewards:0

[step 3] Obs:
Forest Path This is a path winding through a dimly lit forest. The path heads north south here. One particularly large tree with some low branches stands at the edge of the path.

[step 3] Act: climb branches

[step 3] Reward:0, Cumulative Rewards:0

[step 4] Obs:
Up a Tree You are about 10 feet above the ground nestled among some large branches. The nearest branch above you is above your reach. Beside you on the branch is a small bird's nest. In the bird's nest is a large egg encrusted with precious jewels, apparently scavenged by a childless songbird. The egg is covered with fine gold inlay, and ornamented in lapis lazuli and mother of pearl. Unlike most eggs, this one is hinged and closed with a delicate looking clasp. The egg appears extremely fragile.

[step 4] Act: pick up egg

[step 4] Reward:5, Cumulative Rewards:5

[step 5] Obs:
Taken.

[step 5] Act: climb

[step 5] Reward:0, Cumulative Rewards:5

[step 6] Obs:
Forest Path

[step 6] Act: south

[step 6] Reward:0, Cumulative Rewards:5

[step 7] Obs:
North of House You are facing the north side of a white house. There is no door here, and all the windows are boarded up. To the north a narrow path winds through the trees.

[step 7] Act: east

[step 7] Reward:0, Cumulative Rewards:5

[step 8] Obs:
Behind House You are behind the white house. A path leads into the forest to the east. In one corner of the house there is a small window which is slightly ajar.

[step 8] Act: break in window

[step 8] Reward:0, Cumulative Rewards:5

[step 9] Obs:
With great effort, you open the window far enough to allow entry.

[step 9] Act: hatch egg

[step 9] Reward:0, Cumulative Rewards:5

[step 10] Obs:
There is a noticeable crunch from beneath you, and inspection reveals that the egg is lying open, badly damaged. There is a golden clockwork canary nestled in the egg. It seems to have recently had a bad experience. The mountings for its jewel like eyes are empty, and its silver beak is crumpled. Through a cracked crystal window below its left wing you can see the remains of intricate machinery. It is not clear what result winding it would have, as the mainspring seems sprung.

[step 10] Act: break in house

[step 10] Reward:10, Cumulative Rewards:15

[step 11] Obs:
Kitchen You are in the kitchen of the white house. A table seems to have been used recently for the preparation of food. A passage leads to the west and a dark staircase can be seen leading upward. A dark chimney leads down and to the east is a small window which is open. On the table is an elongated brown sack, smelling of hot peppers. A bottle is sitting on the table. The glass bottle contains:   A quantity of water

[step 11] Act: west

[step 11] Reward:0, Cumulative Rewards:15

[step 12] Obs:
Living Room You are in the living room. There is a doorway to the east, a wooden door with strange gothic lettering to the west, which appears to be nailed shut, a trophy case, and a large oriental rug in the center of the room. Above the trophy case hangs an elvish sword of great antiquity. A battery powered brass lantern is on the trophy case.

[step 12] Act: break in case

[step 12] Reward:0, Cumulative Rewards:15

[step 13] Obs:
Opened.

[step 13] Act: east

[step 13] Reward:0, Cumulative Rewards:15

[step 14] Obs:
Kitchen On the table is an elongated brown sack, smelling of hot peppers. A bottle is sitting on the table. The glass bottle contains:   A quantity of water

[step 14] Act: west

[step 14] Reward:0, Cumulative Rewards:15

[step 15] Obs:
Living Room Above the trophy case hangs an elvish sword of great antiquity. A battery powered brass lantern is on the trophy case.

[step 15] Act: apply canary to case

[step 15] Reward:1, Cumulative Rewards:16

[step 16] Obs:
Done.

[step 16] Act: close egg

[step 16] Reward:0, Cumulative Rewards:16

[step 17] Obs:
Closed.

[step 17] Act: apply egg to case

[step 17] Reward:2, Cumulative Rewards:18

[step 18] Obs:
Done.

[step 18] Act: pick up sword

[step 18] Reward:0, Cumulative Rewards:18

[step 19] Obs:
Taken.

[step 19] Act: drop sword

[step 19] Reward:0, Cumulative Rewards:18

[step 20] Obs:
Dropped.

[step 20] Act: activa lantern

[step 20] Reward:0, Cumulative Rewards:18

[step 21] Obs:
The brass lantern is now on.

[step 21] Act: pick up canary

[step 21] Reward:-1, Cumulative Rewards:17

[step 22] Obs:
Taken.

[step 22] Act: pick up lantern

[step 22] Reward:0, Cumulative Rewards:17

[step 23] Obs:
Taken.

[step 23] Act: apply canary to case

[step 23] Reward:1, Cumulative Rewards:18

[step 24] Obs:
Done.

[step 24] Act: pull rug

[step 24] Reward:0, Cumulative Rewards:18

[step 25] Obs:
With a great effort, the rug is moved to one side of the room, revealing the dusty cover of a closed trap door.

[step 25] Act: break in cover

[step 25] Reward:0, Cumulative Rewards:18

[step 26] Obs:
The door reluctantly opens to reveal a rickety staircase descending into darkness.

[step 26] Act: climb

[step 26] Reward:25, Cumulative Rewards:43

[step 27] Obs:
The trap door crashes shut, and you hear someone barring it.  Cellar You are in a dark and damp cellar with a narrow passageway leading north, and a crawlway to the south. On the west is the bottom of a steep metal ramp which is unclimbable.

[step 27] Act: south

[step 27] Reward:0, Cumulative Rewards:43

[step 28] Obs:
East of Chasm You are on the east edge of a chasm, the bottom of which cannot be seen. A narrow passage goes north, and the path you are on continues to the east.

[step 28] Act: east

[step 28] Reward:0, Cumulative Rewards:43

[step 29] Obs:
Gallery This is an art gallery. Most of the paintings have been stolen by vandals with exceptional taste. The vandals left through either the north or west exits. Fortunately, there is still one chance for you to be a vandal, for on the far wall is a painting of unparalleled beauty.

[step 29] Act: pick up art

[step 29] Reward:4, Cumulative Rewards:47

[step 30] Obs:
Taken.

[step 30] Act: north

[step 30] Reward:0, Cumulative Rewards:47

[step 31] Obs:
Studio This appears to have been an artist's studio. The walls and floors are splattered with paints of 69 different colors. Strangely enough, nothing of value is hanging here. At the south end of the room is an open door (also covered with paint). A dark and narrow chimney leads up from a fireplace; although you might be able to get up it, it seems unlikely you could get back down. Loosely attached to a wall is a small piece of paper.

[step 31] Act: climb chimney

[step 31] Reward:0, Cumulative Rewards:47

[step 32] Obs:
Kitchen On the table is an elongated brown sack, smelling of hot peppers. A bottle is sitting on the table. The glass bottle contains:   A quantity of water

[step 32] Act: west

[step 32] Reward:0, Cumulative Rewards:47

[step 33] Obs:
Living Room There is a sword here. Your collection of treasures consists of:     A broken clockwork canary     A broken jewel encrusted egg

[step 33] Act: east

[step 33] Reward:0, Cumulative Rewards:47

[step 34] Obs:
Kitchen On the table is an elongated brown sack, smelling of hot peppers. A bottle is sitting on the table. The glass bottle contains:   A quantity of water

[step 34] Act: west

[step 34] Reward:0, Cumulative Rewards:47

[step 35] Obs:
Living Room There is a sword here. Your collection of treasures consists of:     A broken clockwork canary     A broken jewel encrusted egg

[step 35] Act: east

[step 35] Reward:0, Cumulative Rewards:47

[step 36] Obs:
Kitchen On the table is an elongated brown sack, smelling of hot peppers. A bottle is sitting on the table. The glass bottle contains:   A quantity of water

[step 36] Act: west

[step 36] Reward:0, Cumulative Rewards:47

[step 37] Obs:
Living Room There is a sword here. Your collection of treasures consists of:     A broken clockwork canary     A broken jewel encrusted egg

[step 37] Act: east

[step 37] Reward:0, Cumulative Rewards:47

[step 38] Obs:
Kitchen On the table is an elongated brown sack, smelling of hot peppers. A bottle is sitting on the table. The glass bottle contains:   A quantity of water

[step 38] Act: west

[step 38] Reward:0, Cumulative Rewards:47

[step 39] Obs:
Living Room There is a sword here. Your collection of treasures consists of:     A broken clockwork canary     A broken jewel encrusted egg

[step 39] Act: east

[step 39] Reward:0, Cumulative Rewards:47

[step 40] Obs:
Kitchen On the table is an elongated brown sack, smelling of hot peppers. A bottle is sitting on the table. The glass bottle contains:   A quantity of water

[step 40] Act: west

[step 40] Reward:0, Cumulative Rewards:47

[step 41] Obs:
Living Room There is a sword here. Your collection of treasures consists of:     A broken clockwork canary     A broken jewel encrusted egg

[step 41] Act: east

[step 41] Reward:0, Cumulative Rewards:47

[step 42] Obs:
Kitchen On the table is an elongated brown sack, smelling of hot peppers. A bottle is sitting on the table. The glass bottle contains:   A quantity of water

[step 42] Act: close window

[step 42] Reward:0, Cumulative Rewards:47

[step 43] Obs:
The window closes (more easily than it opened).

[step 43] Act: west

[step 43] Reward:0, Cumulative Rewards:47

[step 44] Obs:
Living Room There is a sword here. Your collection of treasures consists of:     A broken clockwork canary     A broken jewel encrusted egg

[step 44] Act: break in egg

[step 44] Reward:0, Cumulative Rewards:47

[step 45] Obs:
Opened.

[step 45] Act: pick up canary

[step 45] Reward:-1, Cumulative Rewards:46

[step 46] Obs:
Taken.

[step 46] Act: apply canary to case

[step 46] Reward:1, Cumulative Rewards:47

[step 47] Obs:
Done.

[step 47] Act: pick up canary

[step 47] Reward:-1, Cumulative Rewards:46

[step 48] Obs:
Taken.

[step 48] Act: apply canary to case

[step 48] Reward:1, Cumulative Rewards:47

[step 49] Obs:
Done.

[step 49] Act: pick up canary

[step 49] Reward:-1, Cumulative Rewards:46

[step 50] Obs:
Taken.

[step 50] Act: apply canary to case

[step 50] Reward:1, Cumulative Rewards:47

[step 51] Obs:
Done.

[step 51] Act: pick up canary

[step 51] Reward:-1, Cumulative Rewards:46

[step 52] Obs:
Taken.

[step 52] Act: apply canary to case

[step 52] Reward:1, Cumulative Rewards:47

[step 53] Obs:
Done.

[step 53] Act: pick up canary

[step 53] Reward:-1, Cumulative Rewards:46

[step 54] Obs:
Taken.

[step 54] Act: apply canary to case

[step 54] Reward:1, Cumulative Rewards:47

[step 55] Obs:
Done.

[step 55] Act: pick up sword

[step 55] Reward:0, Cumulative Rewards:47

[step 56] Obs:
Taken.

[step 56] Act: apply painting to case

[step 56] Reward:6, Cumulative Rewards:53

[step 57] Obs:
Done.

[step 57] Act: east

[step 57] Reward:0, Cumulative Rewards:53

[step 58] Obs:
Kitchen On the table is an elongated brown sack, smelling of hot peppers. A bottle is sitting on the table. The glass bottle contains:   A quantity of water

[step 58] Act: west

[step 58] Reward:0, Cumulative Rewards:53

[step 59] Obs:
Living Room Your collection of treasures consists of:   A painting   A broken clockwork canary   A broken jewel encrusted egg

[step 59] Act: apply sword to case

[step 59] Reward:0, Cumulative Rewards:53

[step 60] Obs:
Done.

[step 60] Act: apply lantern to case

[step 60] Reward:0, Cumulative Rewards:53

[step 61] Obs:
Done.

[step 61] Act: pick up sword

[step 61] Reward:0, Cumulative Rewards:53

[step 62] Obs:
Taken.

[step 62] Act: apply sword to case

[step 62] Reward:0, Cumulative Rewards:53

[step 63] Obs:
Done.

[step 63] Act: pick up painting

[step 63] Reward:-6, Cumulative Rewards:47

[step 64] Obs:
Taken.

[step 64] Act: pick up lantern

[step 64] Reward:0, Cumulative Rewards:47

[step 65] Obs:
Taken.

[step 65] Act: apply lantern to case

[step 65] Reward:0, Cumulative Rewards:47

[step 66] Obs:
Done.

[step 66] Act: apply painting to case

[step 66] Reward:6, Cumulative Rewards:53

[step 67] Obs:
Done.

[step 67] Act: east

[step 67] Reward:0, Cumulative Rewards:53

[step 68] Obs:
Kitchen On the table is an elongated brown sack, smelling of hot peppers. A bottle is sitting on the table. The glass bottle contains:   A quantity of water

[step 68] Act: pick up bottle

[step 68] Reward:0, Cumulative Rewards:53

[step 69] Obs:
Taken.

[step 69] Act: pick up sack

[step 69] Reward:0, Cumulative Rewards:53

[step 70] Obs:
Taken.

[step 70] Act: put quantity on table

[step 70] Reward:0, Cumulative Rewards:53

[step 71] Obs:
Done.

[step 71] Act: west

[step 71] Reward:0, Cumulative Rewards:53

[step 72] Obs:
Living Room Your collection of treasures consists of:   A painting   A brass lantern (providing light)   A sword   A broken clockwork canary   A broken jewel encrusted egg

[step 72] Act: apply sack to case

[step 72] Reward:0, Cumulative Rewards:53

[step 73] Obs:
Done.

[step 73] Act: pick up sack

[step 73] Reward:0, Cumulative Rewards:53

[step 74] Obs:
Taken.

[step 74] Act: fill case with sack

[step 74] Reward:0, Cumulative Rewards:53

[step 75] Obs:
Done.

[step 75] Act: pick up sack

[step 75] Reward:0, Cumulative Rewards:53

[step 76] Obs:
Taken.

[step 76] Act: pick up painting

[step 76] Reward:-6, Cumulative Rewards:47

[step 77] Obs:
Taken.

[step 77] Act: fill case with sack

[step 77] Reward:0, Cumulative Rewards:47

[step 78] Obs:
Done.

[step 78] Act: apply bottle to case

[step 78] Reward:0, Cumulative Rewards:47

[step 79] Obs:
Done.

[step 79] Act: pick up bottle

[step 79] Reward:0, Cumulative Rewards:47

[step 80] Obs:
Taken.

[step 80] Act: apply painting to case

[step 80] Reward:6, Cumulative Rewards:53

[step 81] Obs:
Done.

[step 81] Act: drop bottle

[step 81] Reward:0, Cumulative Rewards:53

[step 82] Obs:
Dropped.

[step 82] Act: east

[step 82] Reward:0, Cumulative Rewards:53

[step 83] Obs:
Kitchen Sitting on the kitchen table is:    A quantity of water

[step 83] Act: west

[step 83] Reward:0, Cumulative Rewards:53

[step 84] Obs:
Living Room There is a glass bottle here. Your collection of treasures consists of:     A painting     A brown sack     A brass lantern (providing light)     A sword     A broken clockwork canary     A broken jewel encrusted egg

[step 84] Act: pick up sack

[step 84] Reward:0, Cumulative Rewards:53

[step 85] Obs:
Taken.

[step 85] Act: east

[step 85] Reward:0, Cumulative Rewards:53

[step 86] Obs:
Kitchen Sitting on the kitchen table is:    A quantity of water

[step 86] Act: west

[step 86] Reward:0, Cumulative Rewards:53

[step 87] Obs:
Living Room There is a glass bottle here. Your collection of treasures consists of:     A painting     A brass lantern (providing light)     A sword     A broken clockwork canary     A broken jewel encrusted egg

[step 87] Act: apply sack to case

[step 87] Reward:0, Cumulative Rewards:53

[step 88] Obs:
Done.

[step 88] Act: east

[step 88] Reward:0, Cumulative Rewards:53

[step 89] Obs:
Kitchen Sitting on the kitchen table is:    A quantity of water

[step 89] Act: west

[step 89] Reward:0, Cumulative Rewards:53

[step 90] Obs:
Living Room There is a glass bottle here. Your collection of treasures consists of:     A brown sack     A painting     A brass lantern (providing light)     A sword     A broken clockwork canary     A broken jewel encrusted egg

[step 90] Act: apply sack to case

[step 90] Reward:0, Cumulative Rewards:53

[step 91] Obs:
That's easy for you to say since you don't even have the brown sack.

[step 91] Act: break in sack

[step 91] Reward:0, Cumulative Rewards:53

[step 92] Obs:
Opening the brown sack reveals a lunch, and a clove of garlic.

[step 92] Act: apply sack to case

[step 92] Reward:0, Cumulative Rewards:53

[step 93] Obs:
That's easy for you to say since you don't even have the brown sack.

[step 93] Act: break in bottle

[step 93] Reward:0, Cumulative Rewards:53

[step 94] Obs:
Opened.

[step 94] Act: apply sack to case

[step 94] Reward:0, Cumulative Rewards:53

[step 95] Obs:
That's easy for you to say since you don't even have the brown sack.

[step 95] Act: pick up sack

[step 95] Reward:0, Cumulative Rewards:53

[step 96] Obs:
Taken.

[step 96] Act: shake sack

[step 96] Reward:0, Cumulative Rewards:53

[step 97] Obs:
The contents of the brown sack spills to the ground.

[step 97] Act: apply sack to case

[step 97] Reward:0, Cumulative Rewards:53

[step 98] Obs:
Done.

[step 98] Act: pick up sword

[step 98] Reward:0, Cumulative Rewards:53

[step 99] Obs:
Taken.

[step 99] Act: east

[step 99] Reward:0, Cumulative Rewards:53

[step 100] Obs:
Kitchen Sitting on the kitchen table is:    A quantity of water

[step 100] Act: drop sword

[step 100] Reward:0, Cumulative Rewards:53

[step 101] Obs:
Dropped.

\subsection{Ztuu}

[step 0] Obs: 
It is with great hope and expectation I send you on this mission," the Grand Inquisitor intones, his voice booming like distant, rumbling thunderheads. "We have discovered what appears to be a piece of the Great Underground Empire, never before explored. I would have sent Grubald the Bold, but he's busy. As is Matchlick the Mighty. So, 'tis you."  "What of Linklaw the Lucky?" you ask shyly.  "Laid up. I'm afraid 'tis you and none other. A simple, solo quest, one of fun and mirth, mischief and merriment, a jaunt, an excursion, a simple outing. Merely explore, enjoy yourself, and bring back news of what lies there, long ago buried for countless ages."  "And Fretsham the Fortunate?" you ask hopefully.  "A fatality!" the Inquisitor booms. "Here," he says with evaporating patience, "take these and be of good cheer." He thrusts into your hands two things: a brass lantern of dubious quality; and a plastic sword of no antiquity or use whatsoever.  "But what of Kolchack the   "  "Enough! Kolchack is contracted. Remember, who is the boss of you!"  He pauses briefly as if awaiting a reply. Then without warning, he answers his own question. "Me! I am the boss of you! Now begone!"  And with that, you embark on...  Zork: The Undiscovered Underground, a text adventure prequel to Zork Grand Inquisitor Copyright 1997 by Activision Inc. Written by Marc Blank and Mike Berlyn, programmed by Gerry Kevin Wilson. Release 16 / Serial number 970828 / Inform v6.13 Library 6/2 Standard interpreter 1.0  New Excavation You are standing at the entrance of a narrow, waist high tunnel that winds out of sight to the northeast. This, you have been told, is the entrance to the newfound areas of the Great Underground Empire. In your left hand hangs a rather unconvincing replica of the traditional adventurer's brass lantern. In the other is an ersatz Elvish sword of no antiquity whatsoever. Its plastic blade, dull as a bread knife, barely reflects the lantern's dim and uneven glow.

[step 1] Act: blow out brass

[step 1] Reward:0, Cumulative Rewards:0

[step 2] Obs: 
The lantern is now off. You had better hope that it comes back on when you come to your senses.

[step 2] Act: climb up northeast

[step 2] Reward:0, Cumulative Rewards:0

[step 3] Obs: 
As you enter the tunnel, you feel a slight rumble beneath your crepe soled boots. As you turn to look back up to the exit, giant, translucent boulders begin to fall from the walls and ceiling. You duck and cover your head    when the dust clears, you feel your heart sink in your chest. You are trapped!  Darkness It's pitch black. You are likely to be eaten by a grue.

[step 3] Act: hit brass

[step 3] Reward:5, Cumulative Rewards:5

[step 4] Obs: 
Your rough treatment of the lantern was enough to get it lit again. This must be your lucky day.  Tunnel You are in a narrow tunnel whose entrance has been blocked by gigantic, translucent boulders. Your only option is to continue to the northeast, where the tunnel gradually widens.  [Your score has just gone up by five points.]

[step 4] Act: climb up northeast

[step 4] Reward:0, Cumulative Rewards:5

[step 5] Obs: 
Cultural Complex This imposing ante room, the center of what was apparently the cultural center of the GUE, is adorned in the ghastly style of the GUE's "Grotesque Period."  With leering gargoyles, cartoonish friezes depicting long forgotten scenes of GUE history, and primitive statuary of pointy headed personages unknown (perhaps very, very distant progenitors of the Flatheads), the place would have been best left undiscovered. North of here, a large hallway passes under the roughly hewn inscription "Convention Center."  To the east, under a fifty story triumphal arch, a passageway the size of a large city boulevard opens into the Royal Theater. A relatively small and unobtrusive sign (perhaps ten feet high) stands nearby. South, a smaller and more dignified (i.e. post Dimwit) path leads into what is billed as the "Hall of Science."

[step 5] Act: north

[step 5] Reward:0, Cumulative Rewards:5

[step 6] Obs: 
Convention Center Lobby You are standing at the entrance of the GUE Convention Center. On the floor, you can see occasional splotches of paint through the grime and dust. Although the whole cultural center appears to have been long abandoned, loud, low pitched sounds emanate from the door standing open to the north. And while these sounds most closely approximate those made by under lubricated granola fabricators, they nonetheless fill you with dread. Next to the door is a sign, and to the sign's left (northwest to you) is a souvenir stand. The cultural center's lobby lies to the south.

[step 6] Act: northwest

[step 6] Reward:0, Cumulative Rewards:5

[step 7] Obs: 
Souvenir Stand You find yourself standing in front of a souvenir stand that is clearly affiliated with the Convention Center. An odd looking salesman sits behind the counter, his head resting on his hand and his eyes glued to a spot somewhere above and behind you (a quick, frightened glance over your shoulder reveals nothing eye catching; indeed, nothing at all.) A notice on the counter points toward a small, curtained room to the north.  On the counter are four candy bars (a Baby Rune, a Multi Implementeers, a ZM100000 and a Forever Gores), some popcorn, some postcards and a chicken finger.  You can see a salesman here.

[step 7] Act: drop sword

[step 7] Reward:0, Cumulative Rewards:5

[step 8] Obs: 
True, the sword is not a formidable weapon, but it is the only one you've got. Oh well.

[step 8] Act: consult salesman

[step 8] Reward:0, Cumulative Rewards:5

[step 9] Obs: 
Although not much of a conversationalist, the salesman leans forward to listen.  Two things are immediately obvious: 1) He is not a regular user of toothpaste or mouthwash and 2) He is obviously wearing a mask.

[step 9] Act: dig baby with mask

[step 9] Reward:8, Cumulative Rewards:13

[step 10] Obs: 
(first taking the mask) With the salesman's hot, fetid breath threatening to exfoliate multiple layers of facial epidermis, you quickly pull the mask from his face. In hellish harmony, two terrifying screams fill the air: yours, as the first human to see a grue; and his, as prelude to a graphic demonstration of why it is that they aren't seen by light of day (to wit: spontaneous grueish combustion, leaving only a greasy slick on the ground by your feet.) Although you will spend the rest of your life trying to forget what you have just seen, the sight of that sickly glowing fur, fish mouthed face and razor like fingers remains forever etched in your mind. You throw the mask away in horror at what you've just seen.  [Your score has just gone up by eight points.]

[step 10] Act: dig baby with forever

[step 10] Reward:1, Cumulative Rewards:14

[step 11] Obs: 
(first taking the Forever Gores)  [Your score has just gone up by one point.] Digging would achieve nothing here.

[step 11] Act: lower baby

[step 11] Reward:1, Cumulative Rewards:15

[step 12] Obs: 
(first taking the Baby Rune)  [Your score has just gone up by one point.] Dropped.

[step 12] Act: take implementeers from baby

[step 12] Reward:1, Cumulative Rewards:16

[step 13] Obs: 
But it isn't there now.  [Your score has just gone up by one point.]

[step 13] Act: take implementeers from baby

[step 13] Reward:1, Cumulative Rewards:17

[step 14] Obs: 
But it isn't there now.  [Your score has just gone up by one point.]

[step 14] Act: take implementeers from baby

[step 14] Reward:1, Cumulative Rewards:18

[step 15] Obs: 
But it isn't there now.  [Your score has just gone up by one point.]

[step 15] Act: take implementeers from baby

[step 15] Reward:1, Cumulative Rewards:19

[step 16] Obs: 
But it isn't there now.  [Your score has just gone up by one point.]

[step 16] Act: take implementeers from baby

[step 16] Reward:1, Cumulative Rewards:20

[step 17] Obs: 
But it isn't there now.  [Your score has just gone up by one point.]

[step 17] Act: take implementeers from baby

[step 17] Reward:1, Cumulative Rewards:21

[step 18] Obs: 
But it isn't there now.  [Your score has just gone up by one point.]

[step 18] Act: take implementeers from baby

[step 18] Reward:1, Cumulative Rewards:22

[step 19] Obs: 
But it isn't there now.  [Your score has just gone up by one point.]

[step 19] Act: take implementeers from baby

[step 19] Reward:1, Cumulative Rewards:23

[step 20] Obs: 
But it isn't there now.  [Your score has just gone up by one point.]

[step 20] Act: take implementeers from baby

[step 20] Reward:1, Cumulative Rewards:24

[step 21] Obs: 
But it isn't there now.  [Your score has just gone up by one point.]

[step 21] Act: take implementeers from baby

[step 21] Reward:1, Cumulative Rewards:25

[step 22] Obs: 
But it isn't there now.  [Your score has just gone up by one point.]

[step 22] Act: take implementeers from baby

[step 22] Reward:1, Cumulative Rewards:26

[step 23] Obs: 
But it isn't there now.  [Your score has just gone up by one point.]

[step 23] Act: take implementeers from baby

[step 23] Reward:1, Cumulative Rewards:27

[step 24] Obs: 
But it isn't there now.  [Your score has just gone up by one point.]

[step 24] Act: take implementeers from baby

[step 24] Reward:1, Cumulative Rewards:28

[step 25] Obs: 
But it isn't there now.  [Your score has just gone up by one point.]

[step 25] Act: take implementeers from baby

[step 25] Reward:1, Cumulative Rewards:29

[step 26] Obs: 
But it isn't there now.  [Your score has just gone up by one point.]

[step 26] Act: take implementeers from baby

[step 26] Reward:1, Cumulative Rewards:30

[step 27] Obs: 
But it isn't there now.  [Your score has just gone up by one point.]

[step 27] Act: take implementeers from baby

[step 27] Reward:1, Cumulative Rewards:31

[step 28] Obs: 
But it isn't there now.  [Your score has just gone up by one point.]

[step 28] Act: take implementeers from baby

[step 28] Reward:1, Cumulative Rewards:32

[step 29] Obs: 
But it isn't there now.  [Your score has just gone up by one point.]

[step 29] Act: lower sword

[step 29] Reward:0, Cumulative Rewards:32

[step 30] Obs: 
(first taking the cheaply made sword) True, the sword is not a formidable weapon, but it is the only one you've got. Oh well.  The lantern dims a little.

[step 30] Act: take implementeers from baby

[step 30] Reward:1, Cumulative Rewards:33

[step 31] Obs: 
But it isn't there now.  [Your score has just gone up by one point.]

[step 31] Act: take implementeers from baby

[step 31] Reward:1, Cumulative Rewards:34

[step 32] Obs: 
But it isn't there now.  [Your score has just gone up by one point.]

[step 32] Act: take implementeers from baby

[step 32] Reward:1, Cumulative Rewards:35

[step 33] Obs: 
But it isn't there now.  [Your score has just gone up by one point.]

[step 33] Act: take implementeers from baby

[step 33] Reward:1, Cumulative Rewards:36

[step 34] Obs: 
But it isn't there now.  [Your score has just gone up by one point.]

[step 34] Act: take implementeers from baby

[step 34] Reward:1, Cumulative Rewards:37

[step 35] Obs: 
But it isn't there now.  [Your score has just gone up by one point.]

[step 35] Act: lower baby

[step 35] Reward:1, Cumulative Rewards:38

[step 36] Obs: 
(first taking the Baby Rune)  [Your score has just gone up by one point.] Dropped.  The lantern flickers and goes out.  It is now pitch dark in here!

[step 36] Act: hit brass

[step 36] Reward:0, Cumulative Rewards:38

[step 37] Obs: 
Your rough treatment of the lantern was enough to get it lit again. This must be your lucky day.  Souvenir Stand You find yourself standing in front of a souvenir stand that is clearly affiliated with the Convention Center.  A notice on the counter points toward a small, curtained room to the north.  On the counter are two candy bars (a Multi Implementeers and a ZM100000), some popcorn, some postcards and a chicken finger.  You can see a Baby Rune and a cheaply made sword here.

[step 37] Act: take implementeers from baby

[step 37] Reward:1, Cumulative Rewards:39

[step 38] Obs: 
But it isn't there now.  [Your score has just gone up by one point.]

[step 38] Act: take implementeers from baby

[step 38] Reward:1, Cumulative Rewards:40

[step 39] Obs: 
But it isn't there now.  [Your score has just gone up by one point.]

[step 39] Act: take implementeers from baby

[step 39] Reward:1, Cumulative Rewards:41

[step 40] Obs: 
But it isn't there now.  [Your score has just gone up by one point.]

[step 40] Act: take implementeers from baby

[step 40] Reward:1, Cumulative Rewards:42

[step 41] Obs: 
But it isn't there now.  [Your score has just gone up by one point.]

[step 41] Act: take implementeers from baby

[step 41] Reward:1, Cumulative Rewards:43

[step 42] Obs: 
But it isn't there now.  [Your score has just gone up by one point.]

[step 42] Act: take implementeers from baby

[step 42] Reward:1, Cumulative Rewards:44

[step 43] Obs: 
But it isn't there now.  [Your score has just gone up by one point.]

[step 43] Act: take implementeers from baby

[step 43] Reward:1, Cumulative Rewards:45

[step 44] Obs: 
But it isn't there now.  [Your score has just gone up by one point.]

[step 44] Act: take implementeers from baby

[step 44] Reward:1, Cumulative Rewards:46

[step 45] Obs: 
But it isn't there now.  [Your score has just gone up by one point.]

[step 45] Act: take implementeers from baby

[step 45] Reward:1, Cumulative Rewards:47

[step 46] Obs: 
But it isn't there now.  [Your score has just gone up by one point.]

[step 46] Act: take implementeers from baby

[step 46] Reward:1, Cumulative Rewards:48

[step 47] Obs: 
But it isn't there now.  [Your score has just gone up by one point.]

[step 47] Act: take implementeers from baby

[step 47] Reward:1, Cumulative Rewards:49

[step 48] Obs: 
But it isn't there now.  [Your score has just gone up by one point.]

[step 48] Act: take implementeers from baby

[step 48] Reward:1, Cumulative Rewards:50

[step 49] Obs: 
But it isn't there now.  [Your score has just gone up by one point.]

[step 49] Act: take implementeers from baby

[step 49] Reward:1, Cumulative Rewards:51

[step 50] Obs: 
But it isn't there now.  [Your score has just gone up by one point.]

[step 50] Act: take implementeers from baby

[step 50] Reward:1, Cumulative Rewards:52

[step 51] Obs: 
But it isn't there now.  [Your score has just gone up by one point.]

[step 51] Act: take implementeers from baby

[step 51] Reward:1, Cumulative Rewards:53

[step 52] Obs: 
But it isn't there now.  [Your score has just gone up by one point.]

[step 52] Act: take implementeers from baby

[step 52] Reward:1, Cumulative Rewards:54

[step 53] Obs: 
But it isn't there now.  [Your score has just gone up by one point.]

[step 53] Act: take implementeers from baby

[step 53] Reward:1, Cumulative Rewards:55

[step 54] Obs: 
But it isn't there now.  [Your score has just gone up by one point.]

[step 54] Act: take implementeers from baby

[step 54] Reward:1, Cumulative Rewards:56

[step 55] Obs: 
But it isn't there now.  [Your score has just gone up by one point.]

[step 55] Act: take implementeers from baby

[step 55] Reward:1, Cumulative Rewards:57

[step 56] Obs: 
But it isn't there now.  [Your score has just gone up by one point.]

[step 56] Act: take implementeers from baby

[step 56] Reward:1, Cumulative Rewards:58

[step 57] Obs: 
But it isn't there now.  [Your score has just gone up by one point.]

[step 57] Act: take implementeers from baby

[step 57] Reward:1, Cumulative Rewards:59

[step 58] Obs: 
But it isn't there now.  [Your score has just gone up by one point.]

[step 58] Act: take implementeers from baby

[step 58] Reward:1, Cumulative Rewards:60

[step 59] Obs: 
But it isn't there now.  [Your score has just gone up by one point.]

[step 59] Act: take implementeers from baby

[step 59] Reward:1, Cumulative Rewards:61

[step 60] Obs: 
But it isn't there now.  [Your score has just gone up by one point.]

[step 60] Act: take implementeers from baby

[step 60] Reward:1, Cumulative Rewards:62

[step 61] Obs: 
But it isn't there now.  [Your score has just gone up by one point.]

[step 61] Act: take implementeers from baby

[step 61] Reward:1, Cumulative Rewards:63

[step 62] Obs: 
But it isn't there now.  [Your score has just gone up by one point.]

[step 62] Act: take all

[step 62] Reward:1, Cumulative Rewards:64

[step 63] Obs: 
Baby Rune: Taken. cheaply made sword: Taken. counter: That's hardly portable. notice: There's no need to take the notice.  [Your score has just gone up by one point.]

[step 63] Act: take implementeers from baby

[step 63] Reward:1, Cumulative Rewards:65

[step 64] Obs: 
But it isn't there now.  [Your score has just gone up by one point.]

[step 64] Act: take implementeers from baby

[step 64] Reward:1, Cumulative Rewards:66

[step 65] Obs: 
But it isn't there now.  [Your score has just gone up by one point.]

[step 65] Act: take implementeers from baby

[step 65] Reward:1, Cumulative Rewards:67

[step 66] Obs: 
But it isn't there now.  [Your score has just gone up by one point.]

[step 66] Act: take implementeers from baby

[step 66] Reward:1, Cumulative Rewards:68

[step 67] Obs: 
But it isn't there now.  [Your score has just gone up by one point.]

[step 67] Act: take implementeers from baby

[step 67] Reward:1, Cumulative Rewards:69

[step 68] Obs: 
But it isn't there now.  [Your score has just gone up by one point.]

[step 68] Act: take implementeers from baby

[step 68] Reward:1, Cumulative Rewards:70

[step 69] Obs: 
But it isn't there now.  The lantern dims a little.  [Your score has just gone up by one point.]

[step 69] Act: take implementeers from baby

[step 69] Reward:1, Cumulative Rewards:71

[step 70] Obs: 
But it isn't there now.  [Your score has just gone up by one point.]

[step 70] Act: take implementeers from baby

[step 70] Reward:1, Cumulative Rewards:72

[step 71] Obs: 
But it isn't there now.  [Your score has just gone up by one point.]

[step 71] Act: take implementeers from baby

[step 71] Reward:1, Cumulative Rewards:73

[step 72] Obs: 
But it isn't there now.  [Your score has just gone up by one point.]

[step 72] Act: take implementeers from baby

[step 72] Reward:1, Cumulative Rewards:74

[step 73] Obs: 
But it isn't there now.  [Your score has just gone up by one point.]

[step 73] Act: take implementeers from baby

[step 73] Reward:1, Cumulative Rewards:75

[step 74] Obs: 
But it isn't there now.  [Your score has just gone up by one point.]

[step 74] Act: lower implementeers

[step 74] Reward:1, Cumulative Rewards:76

[step 75] Obs: 
(first taking the Multi Implementeers)  [Your score has just gone up by one point.] Dropped.  The lantern flickers and goes out.  It is now pitch dark in here!

[step 75] Act: hit brass

[step 75] Reward:0, Cumulative Rewards:76

[step 76] Obs: 
Your rough treatment of the lantern was enough to get it lit again. This must be your lucky day.  Souvenir Stand You find yourself standing in front of a souvenir stand that is clearly affiliated with the Convention Center.  A notice on the counter points toward a small, curtained room to the north.  On the counter are a ZM100000, some popcorn, some postcards and a chicken finger.  You can see a Multi Implementeers here.

[step 76] Act: lower implementeers

[step 76] Reward:1, Cumulative Rewards:77

[step 77] Obs: 
(first taking the Multi Implementeers)  [Your score has just gone up by one point.] Dropped.

[step 77] Act: lower implementeers

[step 77] Reward:1, Cumulative Rewards:78

[step 78] Obs: 
(first taking the Multi Implementeers)  [Your score has just gone up by one point.] Dropped.

[step 78] Act: lower implementeers

[step 78] Reward:1, Cumulative Rewards:79

[step 79] Obs: 
(first taking the Multi Implementeers)  [Your score has just gone up by one point.] Dropped.

[step 79] Act: lower implementeers

[step 79] Reward:1, Cumulative Rewards:80

[step 80] Obs: 
(first taking the Multi Implementeers)  [Your score has just gone up by one point.] Dropped.

[step 80] Act: lower implementeers

[step 80] Reward:1, Cumulative Rewards:81

[step 81] Obs: 
(first taking the Multi Implementeers)  [Your score has just gone up by one point.] Dropped.

[step 81] Act: lower implementeers

[step 81] Reward:1, Cumulative Rewards:82

[step 82] Obs: 
(first taking the Multi Implementeers)  [Your score has just gone up by one point.] Dropped.

[step 82] Act: lower implementeers

[step 82] Reward:1, Cumulative Rewards:83

[step 83] Obs: 
(first taking the Multi Implementeers)  [Your score has just gone up by one point.] Dropped.

[step 83] Act: lower implementeers

[step 83] Reward:1, Cumulative Rewards:84

[step 84] Obs: 
(first taking the Multi Implementeers)  [Your score has just gone up by one point.] Dropped.

[step 84] Act: lower implementeers

[step 84] Reward:1, Cumulative Rewards:85

[step 85] Obs: 
(first taking the Multi Implementeers)  [Your score has just gone up by one point.] Dropped.

[step 85] Act: lower implementeers

[step 85] Reward:1, Cumulative Rewards:86

[step 86] Obs: 
(first taking the Multi Implementeers)  [Your score has just gone up by one point.] Dropped.

[step 86] Act: lower implementeers

[step 86] Reward:1, Cumulative Rewards:87

[step 87] Obs: 
(first taking the Multi Implementeers)  [Your score has just gone up by one point.] Dropped.

[step 87] Act: lower implementeers

[step 87] Reward:1, Cumulative Rewards:88

[step 88] Obs: 
(first taking the Multi Implementeers)  [Your score has just gone up by one point.] Dropped.

[step 88] Act: examine chicken

[step 88] Reward:0, Cumulative Rewards:88

[step 89] Obs: 
Closer examination of this item reveals it to be a (ugh) human finger. Vaguely, you recall that these are a favorite of grue children.

[step 89] Act: lower implementeers

[step 89] Reward:1, Cumulative Rewards:89

[step 90] Obs: 
(first taking the Multi Implementeers)  [Your score has just gone up by one point.] Dropped.

[step 90] Act: lower implementeers

[step 90] Reward:1, Cumulative Rewards:90

[step 91] Obs: 
(first taking the Multi Implementeers)  [Your score has just gone up by one point.] Dropped.

[step 91] Act: lower implementeers

[step 91] Reward:1, Cumulative Rewards:91

[step 92] Obs: 
(first taking the Multi Implementeers)  [Your score has just gone up by one point.] Dropped.

[step 92] Act: lower implementeers

[step 92] Reward:1, Cumulative Rewards:92

[step 93] Obs: 
(first taking the Multi Implementeers)  [Your score has just gone up by one point.] Dropped.

[step 93] Act: lower implementeers

[step 93] Reward:1, Cumulative Rewards:93

[step 94] Obs: 
(first taking the Multi Implementeers)  [Your score has just gone up by one point.] Dropped.

[step 94] Act: lower implementeers

[step 94] Reward:1, Cumulative Rewards:94

[step 95] Obs: 
(first taking the Multi Implementeers)  [Your score has just gone up by one point.] Dropped.

[step 95] Act: lower implementeers

[step 95] Reward:1, Cumulative Rewards:95

[step 96] Obs: 
(first taking the Multi Implementeers)  [Your score has just gone up by one point.] Dropped.

[step 96] Act: lower implementeers

[step 96] Reward:1, Cumulative Rewards:96

[step 97] Obs: 
(first taking the Multi Implementeers)  [Your score has just gone up by one point.] Dropped.

[step 97] Act: drop baby

[step 97] Reward:0, Cumulative Rewards:96

[step 98] Obs: 
Dropped.

[step 98] Act: take bars from baby

[step 98] Reward:1, Cumulative Rewards:97

[step 99] Obs: 
But it isn't there now.  [Your score has just gone up by one point.]

[step 99] Act: take bars from baby

[step 99] Reward:1, Cumulative Rewards:98

[step 100] Obs: 
But it isn't there now.  [Your score has just gone up by one point.]

[step 100] Act: take bars from baby

[step 100] Reward:1, Cumulative Rewards:99

[step 101] Obs: 
But it isn't there now.  [Your score has just gone up by one point.]

\end{document}